\documentclass[journal,twoside,web]{ieeecolor}

\usepackage{jsen}
\usepackage{cite}
\usepackage{textcomp}
\usepackage{wrapfig}

\usepackage{amsmath,mathtools}  %
\usepackage{amsfonts,amssymb}

\usepackage{xspace}    %
\usepackage{balance}

\usepackage{graphicx}
\usepackage{subcaption}
\graphicspath{{figs/}}

\usepackage{stfloats}%

\usepackage{xcolor}

\usepackage{pgf,tikz,pgfplots}
\usepackage{mathrsfs}
\usetikzlibrary{arrows}
\pgfplotsset{compat=1.15}

\usepackage{grffile}
\pgfplotsset{compat=newest}  %
\usetikzlibrary{plotmarks}
\usetikzlibrary{arrows.meta}
\usepgfplotslibrary{patchplots}

\usepackage{tabularx}

\usepackage[ruled]{algorithm}
\usepackage[noend]{algpseudocode}

\makeatletter
\newcommand{\multiline}[1]{%
  \begin{tabularx}{\dimexpr\linewidth-\ALG@thistlm}[t]{@{}X@{}}
    #1
  \end{tabularx}
}
\makeatother

\usepackage[per-mode=symbol]{siunitx}

\usepackage{cite}  %
\usepackage{array}  %
\usepackage{booktabs}           %

\newtheorem{remark}{Remark}

\newcounter{problem}
{\par\endtrivlist\unskip}

\usepackage[extramath,probability,reviewmark]{truonglatexdefs}

\newcommand{\GPModel}[1][f]{\ensuremath{\GGG_{#1}}}

\newcommand{\hilitediff}[1]{#1}

\def\BibTeX{{\rm B\kern-.05em{\sc i\kern-.025em b}\kern-.08em
    T\kern-.1667em\lower.7ex\hbox{E}\kern-.125emX}}
\definecolor{abstractbg}{rgb}{0.89804,0.94510,0.83137}
\begin{document}
\title{ADMM-based Adaptive Sampling Strategy for Nonholonomic Mobile Robotic Sensor Networks}
\author{Viet-Anh Le, \IEEEmembership{Student, IEEE}, Linh Nguyen, \IEEEmembership{Member, IEEE}, and Truong X. Nghiem, \IEEEmembership{Member, IEEE}
\thanks{V-A.~Le and T.~Nghiem are with the School of Informatics, Computing, and Cyber Systems, Northern Arizona University, Flagstaff, AZ 86011, USA (e-mail: \{vl385,truong.nghiem\}@nau.edu).}
\thanks{L.~Nguyen is with the School of Engineering, Information Technology and Physical Sciences, Federation University Australia, Churchill, VIC 3842, Australia (e-mail: l.nguyen@federation.edu.au).}
\thanks{This work has been submitted to the IEEE Sensors Journal for
possible publication. Copyright may be transferred without notice, after
which this version may no longer be accessible.}
}

\IEEEtitleabstractindextext{%
\fcolorbox{abstractbg}{abstractbg}{%
\begin{minipage}{\textwidth}%
\begin{wrapfigure}[18]{r}{3.1in}%
	\scalebox{.6}{\input{figs/abs_fig.tex}}
    
\end{wrapfigure}%
\begin{abstract}
This paper discusses the adaptive sampling problem in a nonholonomic mobile robotic sensor network for efficiently monitoring a spatial field. 
It is proposed to employ Gaussian process to model a spatial phenomenon and predict it at unmeasured positions, which enables the sampling optimization problem to be formulated by the use of the log determinant of a predicted covariance matrix at next sampling locations. 
The control, movement and nonholonomic dynamics constraints of the mobile sensors are also considered in the adaptive sampling optimization problem.
In order to tackle the nonlinearity and nonconvexity of the objective function in the optimization problem we first exploit the linearized alternating direction method of multipliers (L-ADMM) method that can effectively simplify the objective function, though it is computationally expensive since a nonconvex problem needs to be solved exactly in each iteration. 
We then propose a novel approach called the successive convexified ADMM (SC-ADMM) that sequentially convexify the nonlinear dynamic constraints so that the original optimization problem can be split into convex subproblems. 
It is noted that both the L-ADMM algorithm and our SC-ADMM approach can solve the sampling optimization problem in either a centralized or a distributed manner. 
We validated the proposed approaches in 1000 experiments in a synthetic environment with a real-world dataset, where the obtained results suggest that both the L-ADMM and SC-ADMM techniques can provide good accuracy for the monitoring purpose. 
However, our proposed SC-ADMM approach computationally outperforms the L-ADMM counterpart, demonstrating its better practicality.
\end{abstract}

\end{minipage}}}

\maketitle

\section{Inroduction}
\label{sec:intro}

In applications of monitoring environmental spatial fields, such as exploring ecosystems on land and in ocean, observing chemical concentration and monitoring air pollutants and indoor climates \cite{arampatzis_survey_2005,leonard_collective_2007}, a mobile robotic sensor network (MRSN) incorporated by a machine learning based model representing the spatial phenomenon have been widely used due to its universal capability of monitoring the spatial field, exploring the environment and predicting the field at unobserved locations. 
However, due to the resource constraints in the network including limited numbers of the sensors and robots, communication, memory, computation, power, time and motion dynamics, the fundamental yet challenging problem of how to optimally drive the mobile sensors for efficient monitoring is still not fully solved. In this problem, the sensors are expected to take measurements on the most informative sampling paths so that the prediction uncertainty at unobserved positions of interest is minimal. The problem is also known as \textit{adaptive sampling} (see \cite{nguyen_adaptive_2018,paley_mobile_2020} for reviews).

The adaptive sampling problem has been variously formulated and considered in the literature.
For instance, Xu \textit{et al.} in \cite{xu_adaptive_2011} proposed to minimize the Fisher information matrix based objective function to find the optimal sampling locations for a MRSN.
The authors in \cite{xu_mobile_2011} considered a minimization problem where the cost function is derived from the average of the prediction variances over the prespecified target points in order to obtain the next sampling locations for the mobile sensors.
In \cite{xu_sequential_2011}, the maximum a posterior estimation was exploited to design an adaptive sampling strategy for a MRSN by minimizing the prediction error variances.
Likewise, \cite{viseras_decentralized_2016} proposed that each individual sensing agent in a network determines the sampling path by maximizing the predicted variance at its next location. 
While Tiwari \textit{et al.} in \cite{tiwari_resource-constrained_2016} considered a decentralized multi-robot system in which each robot is allocated and responsible for monitoring a local sensing zone, the work \cite{chen_gaussian_2015} discusses a method to partition a MRSN into several small groups such that each group is independent of finding its locations. 
Nevertheless, none of the aforementioned works have considered the constraints, e.g. dynamics, of a MRSN in the adaptive sampling problem.

Some other works have imposed the movement constraint on a MRSN when designing its sampling scheme. For instance, Marchant \etal in \cite{marchant_bayesian_2012} employed a Bayesian optimization approach to derive a sampling strategy, where the traveled distances of the mobile robots were taken into account to balance a trade-off between exploration and exploitation. 
In \cite{jang_2020_multi}, a problem formulation consisting of an objective function that maximizes both the mean value for exploitation and variance estimate for exploration and subjecting to the constraints for collision avoidance was given. 
Some other works used concepts from the information theory to formulate the adaptive sampling optimization problem \cite{huang_adaptive_2012},\cite{xu_efficient_2013}. 
In our previous studies \cite{nguyen_information-driven_2015,nguyen_adaptive_2016}, the conditional entropy and the posterior variances were utilized to formulate an adaptive sampling problem for a resource-constrained MRSN, while the single-integrator dynamics and the collision avoidance scheme were taken into consideration. 
The optimization problems in those works were efficiently resolved by the grid based greedy algorithm. 
However, to the best of our knowledge that mobile platforms with nonholonomic dynamics are popularly used in a MRSN and that the adaptive sampling problem under the nonholonomic dynamics constraint and the movement constraint (\ie to avoid physical collision among robots) is not yet considered. 
Therefore, in this work we propose to incorporate these two constraints into the adaptive sampling optimization problem in a MRSN so that it can be practically implemented in environmental monitoring applications. 
Furthermore, the single-integrator model proposed in our previous works \cite{nguyen_information-driven_2015,nguyen_adaptive_2016} is not controllably feasible to drive the nonholonomic mobile sensors through the environment. 
It is noted that the grid based greedy algorithm impractically handles the adaptive sampling problem for a nonholonomic MRSN given the movement constraint due to difficulty in defining a search region for a mobile sensor at each moving step.

In this paper we present two approaches to tackle the adaptive sampling problem in a MRSN subject to the both nonholonomic dynamics and movement constraints. 
First, we exploit a state-of-the-art optimization method called the linearized alternating direction method of multipliers (L-ADMM) for nonconvex nonsmooth optimization presented in \cite{liu_linearized_2019} to address the problem. 
Nonetheless, it is discussed that the L-ADMM algorithm is limited by its indicator function that is nondifferentiable, which causes the method to be computationally expensive. 
Thus, we propose a novel approach called the successive convexified ADMM (SC-ADMM) that dexterously exploits both the distributed proximal characteristic of the ADMM paradigm \cite{hong_convergence_2016} and the successive convexification programming (SCP) \cite{mao_successive_2016}. 
It is noted that in this work we employ the non-parametric Gaussian process (GP) model to present a spatial field as the trained GP model can be used to effectively predict the field at unobserved positions of interest. 
Thus, the objective function of the sampling problem, which is formed through the conditional entropy \cite{nguyen_information-driven_2015}, can be represented by the log determinant of a predicted covariance matrix obtained by the trained GP model. 
Nevertheless, it is demonstrated that the objective function is nonconvex and highly complicated, which leads to computational intractability in the optimization. 
In both L-ADMM and SC-ADMM, the first-order approximation is employed to simplify the log determinant objective function. 
Additionally, by the use of the SCP \cite{mao_successive_2016}, the proposed SC-ADMM algorithm can also sequentially convexify the nonlinear dynamic constraints in a small trust region around a nominal solution. 
That is, the nonconvex and highly complicated adaptive sampling problem with the nonholonomic and movement constraints can be transformed into the convex subproblems that can be efficiently addressed by any convex optimization toolboxes.

Furthermore, both the L-ADMM algorithm and the proposed SC-ADMM approach can be implemented in a either centralized or distributed manner. 
In the centralized scenario all the computation is conducted at the central station while in the distributed scenario each individual mobile robot takes charge of its own nonholonomic dynamics, control and movement constraints. 
It is noted that exploiting the parallel computing can significantly reduce the computation time and the SC-ADMM algorithm always computationally outperforms the L-ADMM technique in either the computation paradigm. 
All the proposed approaches in this work were evaluated in the synthetic experiments using the realistic dataset, where the obtained results highly demonstrate their effectiveness in monitoring environmental phenomena. In particular, the SC-ADMM algorithm presents to be attracted by practical implementation in real-time systems.

In summary, the main contributions of this paper are threefold:
\begin{enumerate}
\item An adaptive sampling optimization problem for a resource-constrained MRSN is derived in which the non-holonomic dynamics of sensing robots are taken into account.
\item Two ADMM-based algorithms, L-ADMM presented in \cite{liu_linearized_2019}, and our proposed SC-ADMM, are employed to effectively address the complex and non-convex adaptive sampling optimization problem in continuous domain.
\item Both algorithms allows the computation for solving the optimization problem to be distributed to all sensing agents.
\end{enumerate}

The rest of this paper is organized as follows. 
Section~\ref{sec:mrsn} introduces a nonholonomic MRSN for efficiently monitoring a spatial field, where the adaptive sampling optimization problem subject to the nonholonomic and movement constraints is formulated. 
Section~\ref{sec:pxadmm} then presents how to address the adaptive sampling optimization problem by either the L-ADMM algorithm or the proposed SC-ADMM approach. 
The evaluation of the proposed approaches in the synthetic environment is discussed in Section~\ref{sec:simulation} before the conclusions are drawn in Section~\ref{sec:conclusion}.

\section{Nonholonomic Mobile Sensor Networks for Environmental Monitoring}
\label{sec:mrsn}

In the environmental monitoring applications using a MRSN, it is expected that the robotic sensors adaptively conduct sampling at the most informative positions so that their collective measurements can be utilized in a data-driven model, such as a GP, for efficiently predicting the environmental field at unobserved locations. 
However, when moving on a sampling path, a mobile sensor is constrained not only by its minimum distance to other robots to avoid physical collision but also the nonholonomic dynamic configuration, which limits its motions. In order to design an efficient sampling strategy for the mobile sensors, in this section the adaptive sampling optimization problem in the MRSN is mathematically formulated given those constraints.

\subsection{Nonholonomic Mobile Robotic Sensor Networks}
\label{sec2_1}
Let us consider $M$ networked mobile spatial-field sensors. For the simplicity purpose, we assume that all the sensors are identical and take measurements of an environmental field at discrete time steps while their mobile robots navigate through the environment space.
Given the nonholonomic constraints, the dynamics of a mobile robot $i \in \VVV$, $\VVV = \{ 1,\dots,M \}$, are described by the following kinematic unicycle model,
\begin{equation}
  \begin{split}
    \label{eq:ex-cont-dyn}
    \dot{s}_{x,i} &= \cos(\theta_i) v_i \\
    \dot{s}_{y,i} &= \sin(\theta_{i}) v_i \\
    \dot{\theta}_i &= \omega_i, \\
  \end{split}
\end{equation}
where $(s_{x,i}, s_{y,i})$ is the position vector of the robot on a plane, $\theta_i$ is the heading angle, $v_i$ and $\omega_i$ are the linear and angular velocities, respectively. 
And the nonholonomic dynamics \eqref{eq:ex-cont-dyn} are held by the following constraint,
\begin{equation*}
\dot{s}_{x,i} \sin(\theta_{i}) - \dot{s}_{y,i} \cos(\theta_i) = 0.
\end{equation*}
Let $\Delta T$ denote a sampling time, the discrete model of \eqref{eq:ex-cont-dyn} can be specified as follows,
\begin{equation}
  \begin{split}
    \label{eq:ex-disc-dyn}
    {s}_{x,i,t+1} &= s_{x,i,t} + \Delta T \cos(\theta_{i,t}) v_{i,t} \\
    {s}_{y,i,t+1} &= s_{y,i,t} + \Delta T \sin(\theta_{i,t}) v_{i,t} \\
    {\theta}_{i,t+1} &= \theta_{i,t} + \Delta T \omega_{i,t}. \\
  \end{split}
\end{equation}
where $\mathbf{x}_{i,t} = [s_{x,i,t}, s_{y,i,t}, \theta_{i,t}]^T$, $\mathbf{s}_{i,t} = [s_{x,i,t}, s_{y,i,t}]^T$ and $\mathbf{u}_{i,t} = [v_{i,t}, \omega_{i,t}]^T$ are defined as the state vector, position vector and control input vector of a robot $i$ at time $t$.
The mobile platform's dynamics \eqref{eq:ex-disc-dyn} can be aggregated by
\begin{equation}
\mathbf{x}_{i,t+1} = f_d (\mathbf{x}_{i,t}, \mathbf{u}_{i,t}).
\end{equation}
We denote $\mathbf{x}_{t} = [\mathbf{x}_{i,t}]_{i \in \VVV} \in \RR^{3\times M}$, $\mathbf{u}_{t} = [\mathbf{u}_{i,t}]_{i \in \VVV} \in \RR^{2\times M}$, $\mathbf{s}_{t} = [\mathbf{s}_{i,t}]_{i \in \VVV} \in \RR^{2\times M}$ and $\mathbf{s}_{0:t} = [\mathbf{s}_{k}]_{k = 0, \dots, t} \in \RR^{3\times M(t+1)}$, while assume that a MRSN is operated in a compact 2D Euclidean space of interest $\QQQ \subset \RR^2$, \ie $\mathbf{s}_{i,t} \in \QQQ$ for all $i \in \VVV$.
Moreover, due to the limitation on the robot actuation, the control input is bounded by $\mathbf{u}_{i,t} \in \UUU_i$ where $\UUU_i := \{\mathbf{u} \in \RR^{2}\; | \; \mathbf{u}_{\text{min}} \le\mathbf{u} \le \mathbf{u}_{\text{max}}\}$. 

At time step $t$, a noisy measurement of the environmental field of interest taken by the mobile sensor $i$ at the location $\mathbf{s}_{i,t}$ is modeled as
\begin{equation}
  \label{eq:measurement}
  y_{i,t} = h(\mathbf{s}_{i,t}) + w_i,
\end{equation}
where $h: \RR^2 \rightarrow \RR$ is a latent function of the spatial field at $\mathbf{s}_{i,t}$ while $w_i$ is a Gaussian zero-mean independent and identically distributed noise. 
All the sensor measurements at time step $t$ are denoted by $\mathbf{y}_{t} = [y_{i,t}]_{i \in \VVV} \in \RR^{M}$ and the collective measurements from time step 0 to time step $t$ are denoted by $\mathbf{y}_{0:t} = [\mathbf{y}_{k}]_{k = 0, \dots, t} \in \RR^{M(t+1)}$. 

The measurements of the spatial phenomenon collected after each time step are limited as compared with the infinite number of locations in the space of interest. Thus, in the monitoring applications, it is expected to employ the collective measurements to predict the spatial field at unmeasured positions. Statistically, this paradigm can be obtained by the use of the data-driven GP model $\GPModel[h,t]$ specified by a mean function $m(\mathbf{s}_{i,t}; \theta_t)$ and a covariance function $\text{cov}(\mathbf{s}_{i,t}, \mathbf{s}_{j,t}; \theta_t)$. The hyperparameters $\mathbf{\theta}_t$ can be trained based on the data set $\DDD_t = (\mathbf{s}_{0:t},\; \mathbf{y}_{0:t})$ through optimization such as maximizing the likelihood $\mathbf{\theta}_t^\star = \text{argmax}_{\mathbf{\theta}_t} \text{Pr} (\mathbf{y}_{0:t}\; | \; \mathbf{s}_{0:t},\; \mathbf{\theta}_t)$ \cite{williams_gaussian_2006}.
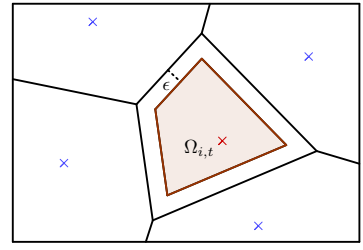
\begin{figure}[!tb]
    \centering
    \scalebox{0.72}{\definecolor{zzttqq}{rgb}{0.6,0.2,0}
\definecolor{uuuuuu}{rgb}{0.26666666666666666,0.26666666666666666,0.26666666666666666}
\definecolor{ududff}{rgb}{0.30196078431372547,0.30196078431372547,1}
\definecolor{ccqqqq}{rgb}{0.8,0,0}
\begin{tikzpicture}[line cap=round,line join=round,>=triangle 45,x=1cm,y=1cm,scale=0.8]
\clip(-0.5,-0.2) rectangle (8.5,5.6);
\fill[line width=1pt,color=zzttqq,fill=zzttqq,fill opacity=0.10000000149011612] (3.2883022418953156,3.06262540009194) -- (4.362184419548067,4.230914802153727) -- (6.312375202780461,2.23537074489267) -- (3.5676746395201575,1.0741512758210026) -- cycle;
\draw [line width=1pt] (0,0)-- (0,5.5);
\draw [line width=1pt] (0,5.5)-- (8,5.5);
\draw [line width=1pt] (8,5.5)-- (8,0);
\draw [line width=1pt] (8,0)-- (0,0);
\draw [line width=1pt] (2.854290532851734,3.177205210829116)-- (3.2309886290369247,0.4960011144521646);
\draw [line width=1pt] (3.2309886290369247,0.4960011144521646)-- (7.015437573887163,2.097114129581112);
\draw [line width=1pt] (7.015437573887163,2.097114129581112)-- (4.35953642939966,4.814780416963676);
\draw [line width=1pt] (4.35953642939966,4.814780416963676)-- (2.854290532851734,3.177205210829116);
\draw [line width=1pt] (0,3.7586347638174336)-- (2.854290532851734,3.177205210829116);
\draw [line width=1pt] (4.35953642939966,4.814780416963676)-- (4.549889877584282,5.5);
\draw [line width=1pt] (7.015437573887163,2.097114129581112)-- (8,1.8044063812772961);
\draw [line width=1pt] (3.2309886290369247,0.4960011144521646)-- (3.0706650364867305,0);
\draw [line width=1pt] (4.362184419548067,4.230914802153727)-- (3.2883022418953156,3.06262540009194);
\draw [line width=1pt] (3.2883022418953156,3.06262540009194)-- (3.5676746395201575,1.0741512758210026);
\draw [line width=1pt] (3.5676746395201575,1.0741512758210026)-- (6.312375202780461,2.23537074489267);
\draw [line width=1pt] (6.312375202780461,2.23537074489267)-- (4.362184419548067,4.230914802153727);
\draw [line width=1pt,dotted] (3.580077412239816,3.966797530163404) -- (3.872412235559546,3.698085722869511);
\draw (3.3,3.6) node[anchor=west] {$\epsilon$};
\draw [line width=1pt,color=zzttqq] (3.2883022418953156,3.06262540009194)-- (4.362184419548067,4.230914802153727);
\draw [line width=1pt,color=zzttqq] (4.362184419548067,4.230914802153727)-- (6.312375202780461,2.23537074489267);
\draw [line width=1pt,color=zzttqq] (6.312375202780461,2.23537074489267)-- (3.5676746395201575,1.0741512758210026);
\draw [line width=1pt,color=zzttqq] (3.5676746395201575,1.0741512758210026)-- (3.2883022418953156,3.06262540009194);
\draw (3.8,2.5) node[anchor=north west] {$\Omega_{i,t}$};
\begin{scriptsize}
\draw [color=ccqqqq] (4.836469513588626,2.3326473210063763)-- ++(-2.5pt,-2.5pt) -- ++(5pt,5pt) ++(-5pt,0) -- ++(5pt,-5pt);
\draw [color=ududff] (1.846096964134431,5.0813736038380055)-- ++(-2.5pt,-2.5pt) -- ++(5pt,5pt) ++(-5pt,0) -- ++(5pt,-5pt);
\draw [color=ududff] (1.1815697309223876,1.8191490044334342)-- ++(-2.5pt,-2.5pt) -- ++(5pt,5pt) ++(-5pt,0) -- ++(5pt,-5pt);
\draw [color=ududff] (5.667128555103679,0.3692714046980693)-- ++(-2.5pt,-2.5pt) -- ++(5pt,5pt) ++(-5pt,0) -- ++(5pt,-5pt);
\draw [color=ududff] (6.830051213224755,4.280920345650773)-- ++(-2.5pt,-2.5pt) -- ++(5pt,5pt) ++(-5pt,0) -- ++(5pt,-5pt);
\end{scriptsize}
\end{tikzpicture}}
    \caption{A constrained movement region $\Omega_{i,t}$ (shaded area) of a mobile sensor $i$ at time $t$ (red cross).} \label{fig:collision}
    \vspace*{-15pt}
\end{figure}

Another constraint considered in the MRSN model in this work is the physical collision avoidance among the mobile sensors when they navigate through the environment. To mathematically formulate this constraint, we employ the Voronoi theory \cite{cortes_coverage_2004}. 
The Voronoi partition of the mobile agent $i$ at time $t$ is defined by
\begin{equation*}
V_{i,t} := \{\mathbf{q} \in \QQQ\; | \; \norm{\mathbf{q} - \mathbf{s}_{i,t}} \le \norm{\mathbf{q} - \mathbf{s}_{j,t}}, \forall j \neq i \}.
\end{equation*}
Moreover, due to geometrical shapes of the robots and the modelling errors caused by the dynamics approximations, we consider the allowable movement region $\Omega_{i,t}$, which is constituted by shrinking the Voronoi cell $V_{i,t}$ by a small safety threshold $\epsilon > 0$, as demonstrated in Fig.~\ref{fig:collision}.

\subsection{Adaptive Sampling Problem for Nonholonomic MRSN}
\begin{figure}[!tb]
    \centering
    \scalebox{.65}{\begin{tikzpicture}[line cap=round,line join=round,>=triangle 45,x=1cm,y=1cm]
\clip(-1.0,-1.0) rectangle (11.5,2.8);
\draw (10.0,-0.1) node[anchor=north] {$t+H$};
\draw (0.0,-0.1) node[anchor=north east] {$t$};
\draw (4.0,-0.1) node[anchor=north] {$t+2$};
\draw (2.0,-0.1) node[anchor=north west] {$t+1$};
\draw (8.0,-0.1) node[anchor=north] {$t+H-1$};
\draw (6.0,-0.1) node[anchor=north] {...};
\draw (-0.8,1.5) node[anchor=north west, align=left] {\large Current \\ \large measurement \\ \large time step};
\draw (9.0,1.5) node[anchor=north west, align=left] {\large Next \\ \large measurement \\ \large time step};
\draw (2.5,2.2) node[anchor=north west, align=left] {\large Time steps for control only,\\ \large no measurement is taken};
\draw [-{latex[scale=3.0]},line width=1pt,dotted] (4.5,1.0) -- (2,0.05);
\draw [-{latex[scale=3.0]},line width=1pt,dotted] (4.5,1.0) -- (4,0.05);
\draw [-{latex[scale=3.0]},line width=1pt,dotted] (4.5,1.0) -- (8,0.05);
\draw [-{latex[scale=3.0]},line width=1pt] (-1.0,0) -- (12.0,0);
\begin{scriptsize}
\draw [color=black] (0,0)-- ++(-2.5pt,0 pt) -- ++(5pt,0 pt) ++(-2.5pt,-2.5pt) -- ++(0 pt,5pt);
\draw [color=black] (10,0)-- ++(-2.5pt,0 pt) -- ++(5pt,0 pt) ++(-2.5pt,-2.5pt) -- ++(0 pt,5pt);
\draw [fill=black] (2,0) circle (1.5pt);
\draw [fill=black] (4,0) circle (1.5pt);
\draw [fill=black] (8,0) circle (1.5pt);
\end{scriptsize}
\end{tikzpicture}}
    \caption{The timeline for our adaptive sampling strategy: assume current measurements have been taken at time step $t$, a sequence of control signals at time step $t$, $t+1$, $\dots$, $t+H-1$ are computed to drive the robots to new positions at time step $t+H$ where new measurements are collected.} \label{fig:sampling_time}
    \vspace*{-15pt}
\end{figure}
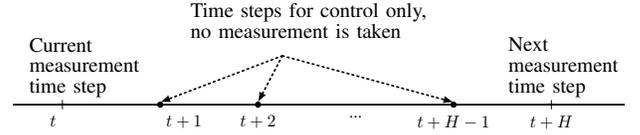

It is defined that at the time step $t$, the mobile sensors take measurements of the environmental phenomenon at their current locations. 
The goal is to find optimal sampling locations of the network at the time $t+H$ and a sequence of control inputs from $t$ to $t+H-1$ using discretized predictive models with control horizon of $H>0$, as can be illustrated by the timeline in Fig.~\ref{fig:sampling_time}. 
Therefore, measurements are taken periodically every $H$ time steps.
Due to the resource constraints in the network including limited numbers of the sensors and robots, communication, memory, computation, power, time and motion dynamics, it is required that the sampling locations lie on the most informative sampling paths. 
To address this problem, we exploit the prediction capability of the GP model $\GPModel[h,t]$ given the collective observations $\mathbf{y}_{0:t}$ to predict the environmental field at unmeasured locations. The predictions can be utilized to optimize the sampling positions. Statistically, the mean vector and covariance matrix of the posterior distribution at the possible next sampling locations $\mathbf{s}_{t+H}$ are given by
\begin{equation*}
\begin{split}
\mathbf{\mu}_{\mathbf{\hat{y}}_{t+H} | \mathbf{y}_{0:t}} &= \mathbf{m}(\mathbf{s}_{t+H}) + \mathbf{\Sigma}_{t+H,0:t} \mathbf{\Sigma}_{0:t}^{-1} \big(\mathbf{y}_{0:t} - \mathbf{m}(\mathbf{s}_{0:t}) \big) \\
\mathbf{\Sigma}_{\mathbf{\hat{y}}_{t+H} | \mathbf{y}_{0:t}} &= \mathbf{\Sigma}_{t+H, t+H} - \mathbf{\Sigma}_{t+H,0:t} \mathbf{\Sigma}_{0:t}^{-1} \mathbf{\Sigma}_{t+H,0:t}^T ,
\end{split}
\end{equation*}
where $\mathbf{m}(\mathbf{s}_{0:t})$ and $\mathbf{m}(\mathbf{s}_{t+H})$ are the mean vectors of the latent variables at $\mathbf{s}_{0:t}$ and $\mathbf{s}_{t+H}$, respectively.
The $M \times M$ matrix $\mathbf{\Sigma}_{t+H, t+H}$ is the covariance matrix at $s_{t+H}$, $\mathbf{\Sigma}_{t+H, 0:t}$ is the cross-covariance matrix between $\mathbf{\hat{y}}_{t+H}$ and $\mathbf{y}_{0:t}$, 
and $\mathbf{\Sigma}_{0:t}$ is the covariance matrix of $\mathbf{y}_{0:t}$. 
For the simplicity purpose, let $\mathbf{\Sigma} (\mathbf{s}_{t+H})$ denote $\mathbf{\Sigma}_{\mathbf{\hat{y}}_{t+H} | \mathbf{y}_{0:t}}$ henceforth.

The main objective in the MRSN for monitoring an environmental phenomenon is minimize the prediction uncertainty at unmeasured positions given the collective measurements. In  \cite{krause_2008_near,nguyen_information-driven_2015} it was proved that minimizing the prediction uncertainty at unobserved locations is equivalent to maximizing the conditional entropy of the spatial field at the next sampling locations. In other words, in the context of the GP model, the optimization criterion for finding the most informative sampling paths can be calculated through the log determinant of the covariance matrix.
The adaptive sampling optimization problem for a general MRSN in a spatial monitoring application can be formulated as follows,
\begin{equation}
\label{eq:min_entropy}
\mathbf{s}_{t+H}^* = \operatorname{argmax}_{s_{t+H}} \operatorname{log\,det} \mathbf{\Sigma} (\mathbf{s}_{t+H}) \text.
\end{equation}
Nonetheless, under the nonholonomic dynamics, movement and control constraints as presented in Section \ref{sec2_1}, the optimization problem \eqref{eq:min_entropy} can be rewritten in a constrained optimization problem format by
\begin{equation}
  \label{eq:coc}
  \begin{split}
    & \begin{multlined}
    \operatorname{minimize}_{\{\mathbf{s}_{t+j+1}, \mathbf{u}_{t+j}\}_{j \in \III_t}} \; f_0(\mathbf{s}_{t+H}) \\
    \quad + \sum_{i=1}^{M} f_i(\{\mathbf{u}_{i,t+j}, \mathbf{s}_{i,t+j+1}\}_{j \in \III_t})
    \end{multlined} \\
    & \text{subject to} \\
    & \quad \mathbf{x}_{i,t+j+1} = f_d (\mathbf{x}_{i,t+j}, \mathbf{u}_{i,t+j}), \\ 
    & \quad \mathbf{u}_{i,t+j} \in \UUU_i,\; \mathbf{s}_{i,t+j+1} \in \Omega_{i,t},  \\
    & \quad \forall j \in \III_t,\; \forall i \in \VVV,  
  \end{split}
\end{equation}
where $f_0(\mathbf{s}_{t+H}) = - \operatorname{log\,det} \big(\mathbf{\Sigma} (\mathbf{s}_{t+H}) \big)$ is the sampling metric,
$f_i(\{\mathbf{u}_{i,t+j}, \mathbf{s}_{i,t+j+1}\}_{j \in \III_t})$ is a convex function of the control cost for each robot with $\III_t = \{0,\dots,H-1\}$. 
For example, in this work $f_i = \sum_{j=0}^{H-1} \norm{\mathbf{u}_{i, t+j}}_{\mathbf{Q}_i}^2 + \norm{\mathbf{u}_{i, t+j} - \mathbf{u}_{i, t+j-1}}_{\mathbf{R}_i}^2 $ that controls both velocity and acceleration of the robot $i$ for smoother movement, 
\hilitediff{where $\mathbf{Q}_i$ and $\mathbf{R}_i$ are positive semidefinite weight matrices with appropriate dimensions.} 
It is noted that the notation $\norm{\nu}_\square^2 = \nu^T \square \nu$ denotes the $\square$-norm of vector $\nu$ with respect to the positive semidefinite matrix $\square$.  

It can be seen that \eqref{eq:coc} is a highly nonconvex and complex optimization problem where the nonconvexity is presented in both the objective and constraint functions. That is, solving the problem by the grid-based methods, \eg \cite{nguyen_information-driven_2015}, may be impractical. Moreover, complexity of the objective function causes \eqref{eq:coc} to be computationally intractable. 
For instance, the popular nonlinear programming solver Ipopt \cite{wachter2006implementation} failed to solve it. Therefore, in this work we employ another algorithm in the optimization domain and propose a new approach to effectively address this highly nonconvex and complex problem.

\vspace{-10pt}
\section{Adaptive Sampling strategy using Distributed Optimization algorithms}
\label{sec:pxadmm}  

\vspace{-10pt}
To solve the adaptive sampling problem \eqref{eq:coc} by distributed optimization algorithms, we first re-write the problem in a splitting form. 
Define $\mathbf{z} = [\mathbf{s}_{i,t+H}^T]_{i \in \VVV}^T \in \RR^{2M}$ and $\mathbf{w}_i = [\mathbf{x}_i^T, \mathbf{u}_i^T]^T \in \RR^{5H}$ where $\mathbf{x}_i = [\mathbf{x}_{i,t+j+1}^T]_{j \in \III_t}^T \in \RR^{3H}$ and $\mathbf{u}_i = [\mathbf{u}_{i,t+j}^T]_{j \in \III_t}^T \in \RR^{2H}$ are the vectors collecting states and control variables of the robot $i$ over a horizon.
The problem \eqref{eq:coc} can be represented in the following splitting form,
\begin{subequations}
  \label{eq:nonholonomic-prb-rwt}
  \begin{align}
    & \underset{\{\mathbf{w}_i\}_{i \in \VVV}, \mathbf{z}}{\operatorname*{minimize}} \; f_0(\mathbf{z}) + \sum_{i=1}^{M} f_i(\mathbf{w}_i) \label{eq:nonholonomic-prb-rwt-obj} \\
    & \text{subject to} \nonumber \\
    & \quad \mathbf{w}_i \in \CC_{i,t}, \forall i \in \VVV \\ 
    & \quad \mathbf{E}_i \mathbf{z} = \mathbf{F}_i \mathbf{w}_i, \; \forall i \in \VVV.
  \end{align}
\end{subequations}
where $\mathbf{E}_i$ and $\mathbf{F}_i$ are transformation matrices that extract $\mathbf{s}_{i,t+H}$ from $\mathbf{z}$ and $\mathbf{w}_i$, respectively.
The nonconvex set $\CC_{i,t}$ represents all the constraints in \eqref{eq:coc} and can be defined by a set of equality constraints $g_{i,j,t} (\mathbf{w}_i) = 0,\; \forall j \in \JJJ_{\text{i,eq,t}}$ and inequality constraints $h_{i,j,t} (\mathbf{w}_i) \le 0, \; \forall j \in \JJJ_{\text{i,ieq,t}}$
where $\JJJ_{\text{i,eq,t}}$ and $\JJJ_{\text{i,ieq,t}}$ are the sets of the equality and inequality constraint indices, respectively.
Based on the definitions of all the constraints in \eqref{eq:coc}, it is noted that for all $i \in \VVV$, $g_{i,j,t} (\mathbf{w}_i), \; \forall j \in \JJJ_{\text{i,eq,t}}$ and $h_{i,j,t} (\mathbf{w}_i), \; \forall j \in \JJJ_{\text{i,ieq,t}}$ are continuously differentiable.

\hilitediff{
The ADMM algorithm, which was presented for convex problems in \cite{boyd2011distributed}, has been demonstrated as an effective method for solving distributed optimization problems.
Some variants of the distributed ADMM algorithm have been developed for nonconvex and nonsmooth optimization in recent years, \eg non-convex ADMM \cite{wang2019global},  linearized ADMM \cite{liu_linearized_2019}, and majorization-ADMM \cite{fan2021spectrally}. 
Since our problem \eqref{eq:nonholonomic-prb-rwt} is constrained, the majorization-ADMM technique in \cite{fan2021spectrally}, which does not handle constraints, is not applicable.
Hence, in this section, we present two approaches based on the distributed ADMM framework with linearization to tackle the non-convex, non-smooth, and highly complex constrained optimization problem \eqref{eq:nonholonomic-prb-rwt}.}
The first method is derived from the work \cite{liu_linearized_2019} where the linearized ADMM (L-ADMM) algorithm is proposed for nonconvex nonsmooth optimization problems. 
Though the L-ADMM technique can find a solution for the problem \eqref{eq:coc}, it is still computationally expensive since a nonconvex optimization problem is involved in each algorithmic iteration. 
Therefore, in the second method, we propose a novel algorithm called successive convexified ADMM (SC-ADMM) that dexterously exploits both the distributed proximal characteristic of the ADMM paradigm \cite{hong_convergence_2016} and the successive convexification programming \cite{mao_successive_2016} to avoid solving non-convex optimization problems. 
The proposed SC-ADMM can address the problem \eqref{eq:coc} more efficiently than the L-ADMM algorithm in terms of computation time, as demonstrated in Section \ref{sec:simulation}.

\subsection{L-ADMM for Adaptive Sampling Problem}
In L-ADMM framework, we minimize the augmented Lagrangian function of the problem \eqref{eq:nonholonomic-prb-rwt} that is defined by
\begin{equation}
\label{eq:aug-lag-nonh-nopen}
\begin{multlined}
L( \mathbf{z}, \{\mathbf{w}_i\}_{i \in \VVV}, \mathbf{\mu} ) = f_0(\mathbf{z}) +  
\sum_{i=1}^M \Big( f_i(\mathbf{w}_i) + I_{\CC_{i,t}}(\mathbf{w}_i) \\
+ (\mathbf{E}_i \mu)^T \left( \mathbf{E}_i \mathbf{z} - \mathbf{F}_i \mathbf{w}_i \right) +
\frac{\rho}{2} \norm{\mathbf{E}_i \mathbf{z} - \mathbf{F}_i  \mathbf{w}_i}_2^2
\Big),
\end{multlined}
\end{equation}
where $\mathbf{\mu} \in \RR^{2M}$ is a vector of the associated dual variables and $\rho$ is a regularization parameter, while $I_{\CC_{i,t}}(\mathbf{w}_i)$ is an indicator function of the set $\CC_{i,t}$, for all $i \in \VVV$.

\begin{algorithm}[!tb]
  \caption{Distributed L-ADMM algorithm}
  \label{alg:ladmm}
  \begin{algorithmic}[1]
    \Require $\mathbf{z}^{(0)}$, $\mathbf{\mu}^{(0)}$, $\epsilon_{res}$, $k_{\mathrm{max}}$, $\rho$, $L$
    \For {$k = 0, \dots, k_{\mathrm{max}}$}
    \State Central station sends the query point $\mathbf{E}_i(\mathbf{z}^{(k)} + \mathbf{\mu}^{(k)}/\rho)$ to agent $i$, $\forall i \in \VVV$
    \State Agent $i$ computes $\mathbf{w}_i^{(k+1)}$ by \eqref{eq:linadmm-1}, $\forall i\in \VVV$, in parallel
    \State Agent $i$ sends $\mathbf{v}_i^{(k)} = \mathbf{F}_i \mathbf{w}_{i}^{(k)}$ to the central station
    \State Central station collects $\mathbf{v}_i^{(k)}$ and forms $\mathbf{v}^{(k)} = [\mathbf{v}_{i}^T]^T_{i\in \VVV}$
    \State Central station computes $\mathbf{z}^{(k+1)}$ by \eqref{eq:linadmm-2} %
    \State Central station updates $\mathbf{\mu}^{(k+1)}$ by \eqref{eq:linadmm-3}
    \If {$ \norm{\mathbf{z}^{(k+1)} - \mathbf{v}^{(k+1)}} < \epsilon_{res}$} 
      \State Stop and return $\mathbf{w}^{(k+1)}$
    \EndIf
    \EndFor
    \State \textbf{return $\mathbf{w}^{(k_{\mathrm{max}})}$} 
  \end{algorithmic}
\end{algorithm}
\setlength{\textfloatsep}{0.1cm}

Given the nonconvex augmented Lagrangian function \eqref{eq:aug-lag-nonh-nopen}, the classical ADMM algorithm \cite{boyd2011distributed} can solve the problem \eqref{eq:nonholonomic-prb-rwt} by performing the following steps.
\begin{subequations}
\label{eq:Cadmm}
  \begin{align}
  & \begin{multlined}
  \mathbf{w}_i^{(k+1)} = \underset{\mathbf{w}_i \in \CC_{i,t}}{\operatorname*{argmin}} \;
    f_i(\mathbf{w}_i) \\
    + \frac{\rho}{2} \norm{\mathbf{F}_i \mathbf{w}_i - \mathbf{E}_i(\mathbf{z}^{(k)} + \frac{\mathbf{\mu}^{(k)}}{\rho})}_2^2, \; \forall i \in \VVV
    \end{multlined} \label{eq:Cadmm-1} \\
  & \mathbf{z}^{(k+1)} = \underset{\mathbf{z}}{\operatorname*{argmin}} \; f_0(\mathbf{z}) +
    \frac{\rho}{2} \norm{\mathbf{z} - \hspace{-2pt} \left( \mathbf{v}^{(k+1)} -  \frac{\mathbf{\mu}^{(k)}}{\rho} \right)}_2^2 \label{eq:Cadmm-2}\\
  & \mathbf{\mu}^{(k+1)} = \mathbf{\mu}^{(k)} + \rho \left(\mathbf{z}^{(k+1)} - \mathbf{v}^{(k+1)} \right), \label{eq:Cadmm-3}
  \end{align}
\end{subequations}
where $\mathbf{v} = [(\mathbf{F}_i \mathbf{w}_{i})^T]^T_{i\in \VVV} \in \RR^{2 M}$.

It can be seen that solving the optimization problem \eqref{eq:Cadmm-2}, which involves the log determinant of a covariance matrix in the function $f_0(\cdot )$, is computationally intractable.
To address this issue, the authors of the work \cite{liu_linearized_2019} proposed in their L-ADMM method to employ the first-order approximation of $f_0$ to find $\mathbf{z}$.
The steps to solving the problem \eqref{eq:nonholonomic-prb-rwt} by the L-ADMM technique are presented as follows,
\begin{subequations}
\label{eq:linadmm}
  \begin{align}
  & \begin{multlined}
  \mathbf{w}_i^{(k+1)} = \underset{\mathbf{w}_i \in \CC_{i,t}}{\operatorname*{argmin}} \;
    f_i(\mathbf{w}_i) \\ 
    + \frac{\rho}{2} \norm{\mathbf{F}_i \mathbf{w}_i - \mathbf{E}_i(\mathbf{z}^{(k)} + \frac{\mathbf{\mu}^{(k)}}{\rho})}_2^2, \; \forall i \in \VVV
    \end{multlined} \label{eq:linadmm-1} \\
  & \mathbf{z}^{(k+1)} = \mathbf{v}^{(k+1)} - \frac{1}{(\rho + L)} \left( \nabla f_0 (\mathbf{v}^{(k+1)}) + \mathbf{\mu}^{(k)} \right) \label{eq:linadmm-2}\\
  & \mathbf{\mu}^{(k+1)} = \mathbf{\mu}^{(k)} + \rho \left(\mathbf{z}^{(k+1)} - \mathbf{v}^{(k+1)} \right), \label{eq:linadmm-3}
  \end{align}
\end{subequations} 
\hilitediff{where $L$ is a Lipschitz constant of $\nabla^T f_0$, \ie $L$ satisfies the following condition
\begin{equation*}
\norm{\nabla^T f_0 (\mathbf{z}) - \nabla^T f_0 (\mathbf{z}')} \le L \norm{\mathbf{z} - \mathbf{z}'}
\end{equation*}
for all $\mathbf{z}, \mathbf{z}' \in \QQQ$.}

To run the L-ADMM algorithm in a distributed manner, at each iteration, each mobile robot is required to solve the optimization problem \eqref{eq:linadmm-1} and then send its correspondingly obtained sampling location $\mathbf{s}_{i,t+H}$ to the central station where the consensus $\mathbf{z}^{(k+1)}$ is computed. The dual variables $\mathbf{\mu}^{(k+1)}$ are then also updated.
The algorithm iterations are repeated until the convergence or the maximum number of the iterations is reached. 
It is noticed that \cite{liu_linearized_2019} provides some criteria to choose the parameters $\rho$ and $L$ so that the L-ADMM algorithm can converge to the Karush-Kuhn-Tucker points of the original nonconvex and nonsmooth problem \eqref{eq:nonholonomic-prb-rwt}. The distributed L-ADMM method is summarized in Algorithm \ref{alg:ladmm}.

\subsection{Successive Convexified ADMM Algorithm}

\begin{algorithm}[!tb]
  \caption{Distributed SC-ADMM algorithm}
  \label{alg:scadmm}
  \begin{algorithmic}[1]
    \Require $\mathbf{z}^{(0)}$, $\mathbf{\mu}^{(0)}$, $\epsilon_{res}$, $k_{\mathrm{max}}$, $\rho$, $L$, $\beta_{\text{fail}}$, $\beta_{\text{succ}}$, $\epsilon_{0}$, $\epsilon_{1}$, $\epsilon_{2}$. %
    \For {$k = 0, \dots, k_{\mathrm{max}}$}
    \State Central station sends the query point $\mathbf{E}_i(\mathbf{z}^{(k)} + \mathbf{\mu}^{(k)}/\rho)$ to agent $i$, $\forall i \in \VVV$
    \State Agent $i$ computes $\mathbf{w}_i^{(k+1)}$ by \eqref{eq:scadmm-1} and \eqref{eq:lin_prbm}, $\forall i \in \VVV$, in parallel
    \State Agent $i$ sends $\mathbf{v}_i^{(k)} = \mathbf{F}_i \mathbf{w}_{i}^{(k)}$ to the central station
    \State Agent $i$ computes $\delta_i^{(k+1)}$ then adjusts the trust region $r_i$ based on Remark~\ref{rmk-rules}
    \State Central station collects $\mathbf{v}_i^{(k)}$ and forms $\mathbf{v}^{(k)} = [\mathbf{v}_{i}]_{i\in \VVV}^T$
    \State Central station computes $\mathbf{z}^{(k+1)}$ by \eqref{eq:scadmm-2} %
    \State Central station updates $\mathbf{\mu}^{(k+1)}$ by \eqref{eq:scadmm-3}
    \If {$ \norm{\mathbf{z}^{(k+1)} - \mathbf{v}^{(k+1)}} < \epsilon_{res}$} 
      \State Stop and return $\mathbf{w}^{(k+1)}$
    \EndIf
    \EndFor
    \State \textbf{return $\mathbf{w}^{(k_{\mathrm{max}})}$} 
  \end{algorithmic}
\end{algorithm}
\setlength{\textfloatsep}{0.1cm}

The main disadvantage of the L-ADMM algorithm \cite{liu_linearized_2019} is that the indicator function $I_{\CC_{i,t}}(\mathbf{w}_i)$ in \eqref{eq:aug-lag-nonh-nopen} is nondifferentiable and hence can not be linearized, leading to the nonconvex optimization problem \eqref{eq:linadmm-1} at each iteration.
Therefore, we propose a novel SC-ADMM method based on the sequential convexification programming \cite{mao_successive_2016} to convexify the non-convex problem \eqref{eq:linadmm-1} by linearizing the nonholonomic dynamics in a small trust region around a nominal solution.
In other words, by the use of the first-order approximations, both the non-linear dynamic constraints and the log determinant of the predicted covariance matrix are linearized. 
The proposed approach can significantly reduces the computation time in solving the nonconvex optimization problem \eqref{eq:linadmm-1} as compared with the L-ADMM algorithm.  
Specifically, instead of considering the indicator function of the constraint sets, we encode the inequality and equality constraints by the exact penalty functions \cite{mao_successive_2016} leading to the nonconvex penalty problem corresponding to \eqref{eq:nonholonomic-prb-rwt} as follows,
\begin{equation}
  \label{eq:nonholonomic-prb-pen}
  \begin{split}
    & \begin{multlined}
    \underset{\mathbf{z}, \{\mathbf{w}_i\}_{i \in \VVV}}{\operatorname*{minimize}} \; J(\mathbf{z}, \{\mathbf{w}_i\}_{i \in \VVV}) = f_0(\mathbf{z}) + 
    \sum_{i=1}^M \Big( 
    f_i(\mathbf{w}_i) + \\
    \sum_{j \in \JJJ_{\text{i,eq}}} \lambda_{i,j} \left | g_{i,j} (\mathbf{w}_i) \right | 
    + \sum_{j \in \JJJ_{\text{i,ieq}}} \tau_{i,j} \operatorname*{max} \left( 0, h_{i,j} (\mathbf{w}_i) \right)
    \Big) 
    \end{multlined} \\
    & \text{subject to} \quad \mathbf{E}_i \mathbf{z} - \mathbf{F}_i \mathbf{w}_i = \mathbf{0}, \; \forall i \in \VVV,
  \end{split}
\end{equation}
where for each mobile sensor $i$, $\lambda_{i,j}, \forall j \in \JJJ_{\text{i,eq}}$ and $\tau_{i,j}, \forall j \in \JJJ_{\text{i,ieq}}$ are the large penalty weights.
The augmented Lagrangian function of the problem \eqref{eq:nonholonomic-prb-pen} can be defined by:
\begin{multline}
\label{eq:aug-lag-nonh}
L( \mathbf{z}, \{\mathbf{w}_i\}_{i \in \VVV}, \mathbf{\mu} ) = f_0(\mathbf{z}) +  
\sum_{i=1}^M \Big( f_i(\mathbf{w}_i) + \\
\sum_{j \in \JJJ_{\text{i,eq}}} \lambda_{i,j} \left | g_{i,j} (\mathbf{w}_i) \right | + 
\sum_{j \in \JJJ_{\text{i,ieq}}} \tau_{i,j} \text{max} \left( 0, h_{i,j} (\mathbf{w}_i) \right) \\
+ (\mathbf{E}_i \mu)^T \left( \mathbf{E}_i \mathbf{z} - \mathbf{F}_i \mathbf{w}_i \right) +
\frac{\rho}{2} \norm{\mathbf{E}_i \mathbf{z} - \mathbf{F}_i  \mathbf{w}_i}_2^2
\Big).
\end{multline}

In the proposed algorithm, we employ the first-order approximation of the non-linear terms in the w-minimization steps to form the convex subproblems.
It is noted that the linearization is computed by the $| \cdot  |$ and $\operatorname*{max}(0, \cdot)$ functions, leading to the following iterative computation steps,
\begin{subequations}
\label{eq:scadmm}
  \begin{align}
  & \mathbf{w}_i^{(k+1)} = \mathbf{w}_i^{(k)} + \mathbf{\Delta}_i^{(k+1)}, \; \forall i \in \VVV \label{eq:scadmm-1} \\
  & \mathbf{z}^{(k+1)} = \mathbf{v}^{(k+1)} - \frac{1}{(\rho + L)} \left( \nabla f_0 (\mathbf{v}^{(k+1)}) + \mathbf{\mu}^{(k)} \right) \label{eq:scadmm-2}\\
  & \mathbf{\mu}^{(k+1)} = \mathbf{\mu}^{(k)} + \rho \left(\mathbf{z}^{(k+1)} - \mathbf{v}^{(k+1)} \right) \label{eq:scadmm-3}
  \end{align}
\end{subequations}
$\mathbf{\Delta}_i^{(k+1)}$ in \eqref{eq:scadmm-1} can be determined by solving the following convex optimization subproblem.
\begin{equation}
\label{eq:lin_prbm}
\begin{split}
  & \begin{multlined}
  \mathbf{\Delta}_i^{(k+1)} = \underset{\mathbf{\Delta}_i}{\operatorname*{argmin}}\; f_i(\mathbf{w}_i^{(k)} + \mathbf{\Delta}_i) \\ 
  + \sum_{j \in \JJJ_{\text{i,eq}}} \lambda_{i,j} \left | g_{i,j} (\mathbf{w}_i^{(k)}) + \nabla g_{i,j} (\mathbf{w}_i^{(k)})^T \mathbf{\Delta}_i \right | \\
  + \sum_{j \in \JJJ_{\text{i,ieq}}} \tau_{i,j} \operatorname*{max} \left( 0, h_{i,j} (\mathbf{w}_i^{(k)}) + \nabla h_{i,j} (\mathbf{w}_i^{(k)})^T \mathbf{\Delta}_i \right) \\
  + \norm{\mathbf{F}_i (\mathbf{w}_i^{(k)} + \mathbf{\Delta}_i) - \mathbf{E}_i \left( \mathbf{z}^{(k)} + \mathbf{\mu}^{(k)}/\rho \right)}^2 
  \end{multlined} \\
  & \text{subject to} \quad \norm{\mathbf{\Delta}_i} \le r_i,
  \end{split}
\end{equation}
where $r_i$ is a small trust-region radius that specifies the local neighborhood around the nominal solution in which the convex optimization subproblem \eqref{eq:lin_prbm} is valid. 
This radius can be adapted by comparing the actual cost value and the predicted cost value, which are respectively given by
\begin{equation*}
\begin{multlined}
J_i(\mathbf{w}_i^{(k+1)}) = f_i(\mathbf{w}_i^{(k+1)}) +
\sum_{j \in \JJJ_{\text{i,eq}}} \lambda_{i,j} \left | g_{i,j} (\mathbf{w}_i^{(k+1)}) \right | \\
+ \sum_{j \in \JJJ_{\text{i,ieq}}} \tau_{i,j} \text{max} \left( 0, h_{i,j} (\mathbf{w}_i^{(k+1)}) \right)
\end{multlined}
\end{equation*}
and 
\begin{equation*}
\begin{multlined}
\tilde{J}_i(\mathbf{\Delta}_i^{(k+1)}) = f_i(\mathbf{w}_i^{(k)}+\mathbf{\Delta}_i^{(k+1)}) \\
+ \sum_{j \in \JJJ_{\text{i,eq}}} \lambda_{i,j} \left | g_{i,j} (\mathbf{w}_i^{(k)}) + \nabla g_{i,j} (\mathbf{w}_i^{(k)})^T \mathbf{\Delta}_i^{(k+1)} \right | \\
+ \sum_{j \in \JJJ_{\text{i,ieq}}} \tau_{i,j} \text{max} \left( 0, h_{i,j} (\mathbf{w}_i^{(k)}) + \nabla h_{i,j} (\mathbf{w}_i^{(k)})^T \mathbf{\Delta}_i^{(k+1)} \right).
\end{multlined}
\end{equation*} 

\begin{remark}
\label{rmk-rules}
(The adjustment rule for the trust-region radius\footnote{This rule is an adapted version of the original adjustment rule in \cite{mao_successive_2016}.})
We compare the difference between the actual cost and the predicted cost, \ie $\delta_i^{(k+1)} = J_i(\mathbf{w}_i^{(k+1)}) - \tilde{J}_i(\mathbf{\Delta}_i^{(k+1)})$, with some predefined thresholds $0 < \epsilon_0 < \epsilon_1 < \epsilon_1 < + \infty$ to adjust the trust region $r_i$ according to the following rule.
\begin{itemize} 
\item If $\delta_i^{(k+1)} > \epsilon_2$, the approximation is considered highly inaccurate, then the solution is rejected and $r_i$ is contracted by a predefined factor $\beta_{\text{fail}} < 1$.
\item If $\epsilon_2 > \delta_i^{(k+1)} > \epsilon_1$, the approximation is considered inaccurate but acceptable, then the solution is accepted. 
However, $r_i$ is still contracted by $\beta_{\text{fail}}$.
\item If $\epsilon_1 > \delta_i^{(k+1)} > \epsilon_0$, the approximation is sufficiently accurate, then the solution is acceptted and $r_i$ is maintained.
\item If $\delta_i^{(k+1)} < \epsilon_0$, the approximation is accurate, then the solution is accepted and $r_i$ is enlarged by a predefined factor $\beta_{\text{succ}} > 1$.
\end{itemize}
\end{remark}

The SC-ADMM approach can be computed in a distributed fashion where each individual robot calculates its own nonholonomic dynamics, control and movement constraints before sending the results to the central station. The approximated linearization of both the objective and constraint functions and the parallel computing allows the SC-ADMM method to significantly reduce its computation time. In other words, the proposed algorithm is highly practically scalable, particularly in a large-scale network.

The distributed SC-ADMM approach for solving the nonholonomic adaptive sampling optimization problem \eqref{eq:coc} is summarized in Algorithm~\ref{alg:scadmm}, where the adjustment rule is applied for computing the trust-region radius. 
Although the convergence analysis of SC-ADMM algorithm have not been derived in literature, the algorithm can be converged in practice \hilitediff{as illustrated in Section~\ref{sec:simulation}}.

\begin{remark}
In the algorithm design, we assume that the convex constraints are also linearized to facilitate the formulation and design. 
However, in practice, they are handled explicitly (in their original form) by the convex optimization solver.
\end{remark}

\section{Simulation Results and Discussion}
\label{sec:simulation}

\begin{figure}[!tb]
    \centering
    \scalebox{0.9}{\input{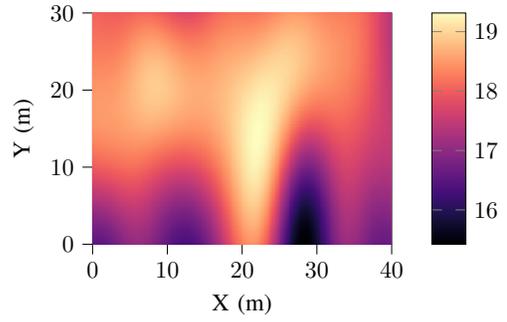}}
    \caption{A generated ground truth of the indoor temperature field.} \label{fig:exact_field}
    \vspace{-5pt}
\end{figure}

\begin{figure*}[!tb]
\centering
\begin{subfigure}{.32\textwidth}
\centering
\scalebox{0.6}{\begin{tikzpicture}

\begin{axis}[
width=8.0cm,
height=5.0cm,
legend cell align={left},
legend style={fill opacity=0.8, draw opacity=1, text opacity=1, draw=white!80!black},
tick align=outside,
tick pos=left,
x grid style={white!69.0196078431373!black},
xlabel={Iterations},
xmajorgrids,
xmin=-1.0, xmax=45.0,
xtick={0,10,20,30,40},
xtick style={color=black},
y grid style={white!69.0196078431373!black},
ylabel={Amplitude},
ymajorgrids,
ymin=-0.1, ymax=2.5,
ytick={0,0.5,1.0,1.5,2.0,2.5},
ytick style={color=black}
]
\addplot [semithick, blue]
table {%
0 2.2703971614648
1 1.68745269826023
2 0.803185707753682
3 0.41763678991536
4 0.416832507103365
5 0.364574087019626
6 0.317765360302326
7 0.317774146044041
8 0.284693023717075
9 0.294976006887717
10 0.260826532955599
11 0.272128887535443
12 0.216658496717304
13 0.230769053751763
14 0.136229359110883
15 0.176176706230862
16 0.0577595796252695
17 0.124463013889601
18 0.0543894669657679
19 0.135374071793741
20 0.211700494476961
21 0.326054500815696
22 0.350053833347241
23 0.254443198572911
24 0.213711562668573
25 0.200365164258054
26 0.180230671792415
27 0.15126711653382
28 0.137044364248181
29 0.105275436495995
30 0.0691775008881983
31 0.0532423060477214
32 0.0405019296628737
33 0.0310611250282342
34 0.0218786651193952
35 0.0167298414346178
36 0.0141790456417787
37 0.0125465330127028
38 0.0113335155576486
39 0.0102949541000612
40 0.00622326649646965
41 0.00397169913265216
42 0.00239781450724053
43 0.00160354600198787
44 0.00102885172226408
};
\end{axis}

\end{tikzpicture}}
\caption{Residual convergence} \label{fig:residual}
\end{subfigure}
\begin{subfigure}{.32\textwidth}
\centering
\scalebox{0.6}{\begin{tikzpicture}

\begin{axis}[
width=8.0cm,
height=5.0cm,
legend cell align={left},
legend style={fill opacity=0.8, draw opacity=1, text opacity=1, draw=white!80!black},
tick align=outside,
tick pos=left,
x grid style={white!69.0196078431373!black},
xlabel={Iterations},
xmajorgrids,
xmin=-1.0, xmax=45.0,
xtick={0,10,20,30,40},
xtick style={color=black},
y grid style={white!69.0196078431373!black},
ylabel={Amplitude},
ymajorgrids,
ymin=10.0, ymax=251.0,
ytick={50,100,150,200,250},
ytick style={color=black}
]
\addplot [semithick, blue]
table {%
0 231.199850414494
1 145.148828075045
2 82.0508879937618
3 64.8252267431847
4 47.7295785771079
5 45.4548376934487
6 38.0264250176325
7 38.2238919301818
8 33.3570702600161
9 34.4023163278576
10 30.6230001419261
11 32.0500229421265
12 28.9328696772292
13 30.3505109255608
14 27.9180925662822
15 28.904649189652
16 27.2851125493242
17 27.68450526587
18 26.7185138026863
19 26.5413905973
20 25.8043144981212
21 24.9651453608846
22 24.2118044392486
23 24.1159862680329
24 24.080936106761
25 24.0742009985631
26 24.1701476424551
27 24.3397625598839
28 24.4793977831833
29 24.6536050875349
30 24.7807951123522
31 24.8325909085147
32 24.8454755577192
33 24.8444940413332
34 24.8501565545986
35 24.8625121958008
36 24.8689455589092
37 24.8744839799287
38 24.8831980945207
39 24.8958927046716
40 24.9145801186702
41 24.9244768444732
42 24.9375367079867
43 24.9402234281952
44 24.9473031251648
};
\end{axis}

\end{tikzpicture}}
\caption{Objective convergence} \label{fig:obj}
\end{subfigure}
\begin{subfigure}{.32\textwidth}
\centering
\scalebox{0.6}{\input{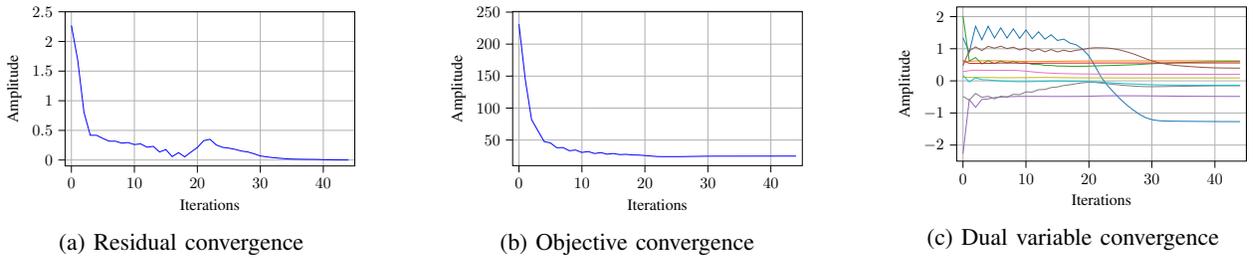}}
\caption{Dual variable convergence} \label{fig:dual}
\end{subfigure}
\caption{\hilitediff{Convergence properties of the SC-ADMM algorithm in a specific time step.}}
\label{fig:convergence}
\vspace{-5pt}
\end{figure*}

To validate effectiveness of the proposed approach, we conducted several experiments in a synthetic environment using the Intel Berkeley Research Lab's temperature data \cite{intellabdata2004}.
The synthetic experiments were executed on a DELL computer with a \SI{3.0}{GHz} Intel Core i5 CPU and \SI{8}{Gb} RAM, where the Python programming language was employed as a platform to implement the algorithms. 
For the GP model, it was proposed to use the constant mean and the squared exponential covariance functions.
More importantly, to present an environmental field in the experiments, a ground truth GP model of an indoor spatial temperature phenomenon was trained based on the 54 temperature measurements gathered by the 54 sensors in the Intel Berkeley Research Lab. 
It is noted that the ground truth model, as demonstrated in Fig. \ref{fig:exact_field}, was utilized for the two purposes: (1) generating sensor measurements of the temperature field and (2) verifying predictions.

In all the synthetic experiments, 5 networked mobile sensors were expected to efficiently monitor the indoor temperature field in an environment of size \SI{40}{m}-by-\SI{30}{m}.
The linear velocity for each mobile sensor was bounded between $v_{\text{min}} = -2$ (\si{m/s}) and $v_{\text{max}} = 2$ (\si{m/s}) while its angular velocity was bounded between $\omega_{\text{min}} = -\pi$ (\si{rad/s}) and $\omega_{\text{max}} = \pi$ (\si{rad/s}). 
The parameters in the control cost functions were chosen by $\mathbf{Q}_i = \operatorname*{diag} ([0.01,\; 0.01])$ and $\mathbf{R}_i = \operatorname*{diag} ([1.0,\; 1.0])$.  
Moreover the sampling time and the length of the control horizon were set to $\Delta T = \SI{0.2}{s}$ and $H = 10$, respectively.
It is noticed that the initial locations of the mobile sensors were arbitrarily chosen at each experiment.
More importantly, in order to address the adaptive sampling problem \eqref{eq:coc}, both the L-ADMM and SC-ADMM algorithms were implemented. 
While some common parameters between the two algorithms were set to $\rho = 0.1$, 
$L = 0.01$,
$\epsilon_{res} = 10^{-3}$ and
$k_{\text{max}} = 100$,
the specific parameters for the SC-ADMM approach were set to 
$\lambda_{i,j} = \tau_{i,j} = 10^{6}$,
$\beta_{\text{fail}} = 0.5$,
$\beta_{\text{succ}} = 2.0$,
$\epsilon_{0} = 1$,
$\epsilon_{1} = 10^{2}$,
$\epsilon_{2} = 10^{3}$,
$r_{\text{min}} = 10^{-6}$ and
$r_{\text{max}} = 1.0$, respectively.
\hilitediff{The convergence properties of the SC-ADMM algorithm in the first time step of a specific experiment are demonstrated in Fig.~\ref{fig:convergence}. 
While Fig.~\ref{fig:residual} shows the norm of the residuals $\norm{\mathbf{\epsilon}^{(k)}} = \norm{\mathbf{z}^{(k)} - \mathbf{v}^{(k)}}$, Figures~\ref{fig:obj} and \ref{fig:dual} illustrate the objective values $J(\mathbf{z}^{(k)}, \{\mathbf{w}_i^{(k)}\}_{i \in \VVV})$ and the values of all elements in the vector of the dual variables $\mathbf{\mu}^{(k)}$, respectively.}

\begin{figure*}[!tb]
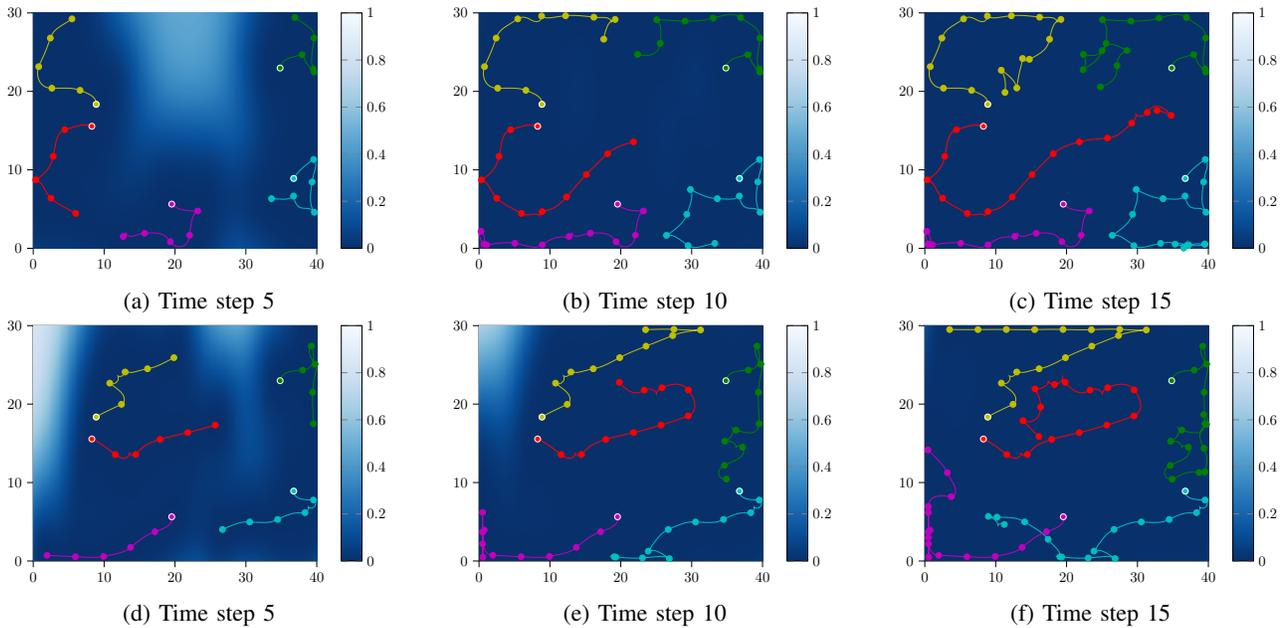

\centering
\begin{subfigure}{.32\textwidth}
\centering
\scalebox{.55}{\input{figs/scadmm_var_5.tex}}
\caption{Time step 5}
\end{subfigure}
\begin{subfigure}{.32\textwidth}
\centering
\scalebox{.55}{\input{figs/scadmm_var_10.tex}}
\caption{Time step 10}  
\end{subfigure}
\begin{subfigure}{.32\textwidth}
\centering
\scalebox{.55}{\input{figs/scadmm_var_15.tex}}
\caption{Time step 15}  
\end{subfigure}

\begin{subfigure}{.32\textwidth}
\centering
\scalebox{.55}{\input{figs/ladmm_var_5.tex}}
\caption{Time step 5}
\end{subfigure}
\begin{subfigure}{.32\textwidth}
\centering
\scalebox{.55}{\input{figs/ladmm_var_10.tex}}
\caption{Time step 10}  
\end{subfigure}
\begin{subfigure}{.32\textwidth}
\centering
\scalebox{.55}{\input{figs/ladmm_var_15.tex}}
\caption{Time step 15}  
\end{subfigure}
\caption{An example of the predicted variances in the entire environment and the sampling trajectories of the mobile sensors obtained by the SC-ADMM algorithm ((a), (b) and (c)) and L-ADMM algorithm ((d), (e) and (f)) at some specific time instants. The starting locations of the mobile robots are shown in the white circles.}
\label{fig:centralized_colorbar}
\vspace{-20pt}
\end{figure*}

It is understood that at the beginning of the monitoring process, all the mobile sensors have no information about the spatial temperature phenomenon. 
That is, they can start at any random locations. 
To demonstrate that our proposed algorithm is always valid, we conducted the 1000 experiments given the arbitrary starting positions of the robotic sensors. 
Overall, the obtained results show that the network of the 5 mobile sensors efficiently monitored the temperature field. 
In other words, the proposed approach drove the robotic sensors on the most informative sampling paths while the prediction uncertainty of the spatial phenomenon in the entire environment was significantly reduced after every sampling step where the sensors took measurements. 
For instance, the sampling trajectories of the mobile sensors at one random experiment example are illustrated in Fig. \ref{fig:centralized_colorbar}. 
These sampling paths were obtained by both the L-ADMM algorithm and our proposed SC-ADMM technique. 
It is noted that the trajectories are plotted on the background with the predicted variances of the spatial temperature field in the entire environment in order to highlight how considerably the robot navigations reduced the prediction uncertainty of the temperature. 
Given the color bars, it can be seen that at each sampling step each robotic sensor tended to move to the location with the highest prediction uncertainty in its allowable movement region so that the predicted variance was minimized.

\begin{figure*}[!tb]
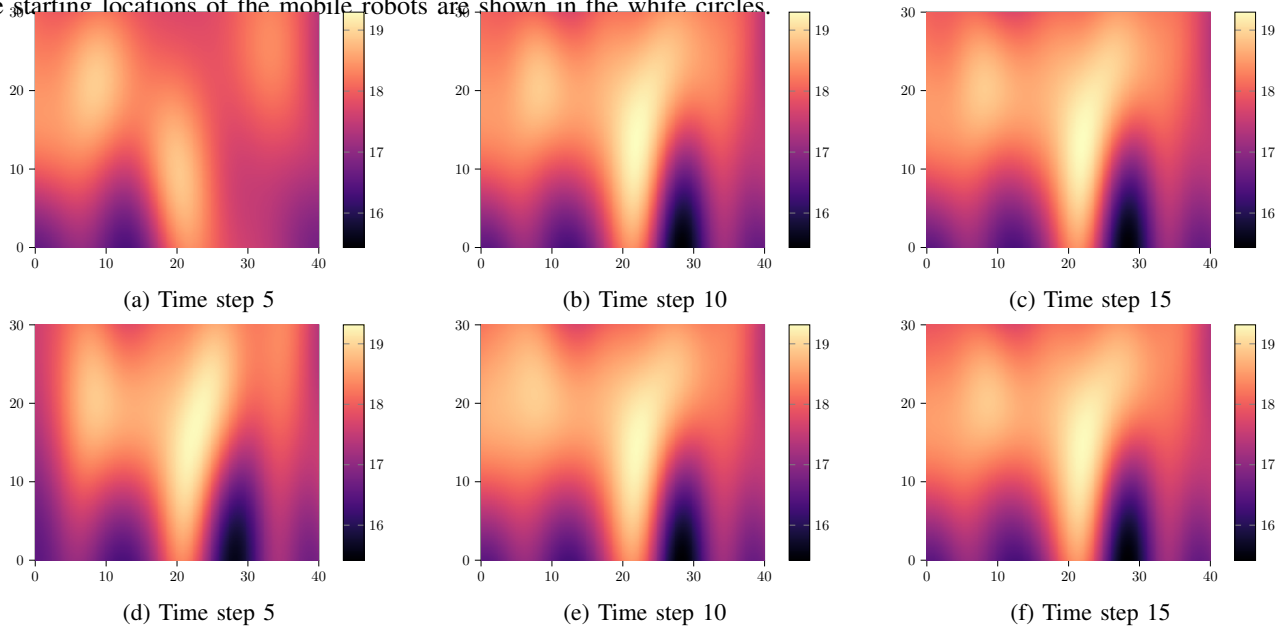

\centering
\begin{subfigure}{.32\textwidth}
\centering
\scalebox{.55}{\input{figs/scadmm_mean_5.tex}}
\caption{Time step 5}
\end{subfigure}
\begin{subfigure}{.32\textwidth}
\centering
\scalebox{.55}{\input{figs/scadmm_mean_10.tex}}
\caption{Time step 10}
\end{subfigure}
\begin{subfigure}{.32\textwidth}
\centering
\scalebox{.55}{\input{figs/scadmm_mean_15.tex}}
\caption{Time step 15}  \label{scadmm_mean_15}
\end{subfigure}

\centering
\begin{subfigure}{.32\textwidth}
\centering
\scalebox{.55}{\input{figs/ladmm_mean_5.tex}}
\caption{Time step 5}
\end{subfigure}
\begin{subfigure}{.32\textwidth}
\centering
\scalebox{.55}{\input{figs/ladmm_mean_10.tex}}
\caption{Time step 10}
\end{subfigure}
\begin{subfigure}{.32\textwidth}
\centering
\scalebox{.55}{\input{figs/ladmm_mean_15.tex}}
\caption{Time step 15}  \label{ladmm_mean_15}
\end{subfigure}
\caption{An example of the predicted means in the entire environment obtained by the SC-ADMM algorithm ((a), (b) and (c)) and L-ADMM algorithm ((d), (e) and (f)) at some specific time instants.}
\label{fig:centralized_field}
\vspace{-20pt}
\end{figure*} 

Furthermore, the predicted temperature fields in the whole environment at the 3 particular time steps of 5, 10 and 15, respectively, in the demonstrated example are depicted in Fig. \ref{fig:centralized_field}. 
Given the sampling trajectories shown in Fig. \ref{fig:centralized_colorbar}, over time it can be seen that the corresponding predicted fields in Fig. \ref{fig:centralized_field} were gradually approaching to the ground truth presented in Fig. \ref{fig:exact_field}. 
Though at the time step of 5 the predicted temperature obtained by the L-ADMM method is slightly better than that obtained by our proposed SC-ADMM algorithm as compared with the ground truth, at the time steps of 10 and 15, differentiation of the predicted temperature results obtained by both the techniques is hardly seen in Fig. \ref{fig:centralized_field}. 
Therefore, we further summarized the results at every sampling step for the 3 other different validation metrics including the average logarithm of predicted variances (ALPVs), the root mean squared errors (RMSEs) and the maximum absolute errors (MAEs). 
These results were computed against the ground truth in Fig. \ref{fig:exact_field} and are demonstrated in Fig. \ref{fig:metrics}. 
It can be clearly seen that from the time step of 8 onwards the proposed SC-ADMM algorithm outperforms the L-ADMM method though at the time step of 15 this outperformance is trivial.
It can be explained by the fact that at the time step of 15 the two sampling networks reached to a large number of 80 measurements, leading to the predicted temperature fields obtained by both the algorithms as demonstrated in Figures \ref{scadmm_mean_15} and \ref{ladmm_mean_15}, respectively, to be highly comparable with the ground truth.

\begin{figure}[!tb]
    \centering
    \scalebox{0.9}{\begin{tikzpicture}

\definecolor{color0}{rgb}{1,0.498039215686275,0.0549019607843137}
\definecolor{color1}{rgb}{1,0.752941176470588,0.796078431372549}
\definecolor{color2}{rgb}{0.67843137254902,0.847058823529412,0.901960784313726}

\begin{axis}[
width=8.8cm,
height=6.0cm,
legend cell align={left},
legend style={fill opacity=0.8, draw opacity=1, text opacity=1, at={(0.03,0.97)}, anchor=north west, draw=white!80!black},
tick align=outside,
tick pos=left,
x grid style={white!69.0196078431373!black},
xlabel={Methods},
xmin=0.5, xmax=5.5,
xtick style={color=black},
xtick={1.5,4.5},
xticklabels={SC-ADMM,L-ADMM},
y grid style={white!69.0196078431373!black},
ylabel={Computation times (s)},
ymin=-0.1, ymax=4.5,
ytick style={color=black}
]
\addplot [black, forget plot]
table {%
1 0.615392239888509
1 0.546703100204468
};
\addplot [black, forget plot]
table {%
1 0.664434957504273
1 0.736371930440267
};
\addplot [black, forget plot]
table {%
0.875 0.546703100204468
1.125 0.546703100204468
};
\addplot [black, forget plot]
table {%
0.875 0.736371930440267
1.125 0.736371930440267
};
\addplot [black, mark=*, mark size=2, mark options={solid,fill opacity=0}, only marks, forget plot]
table {%
1 0.541803073883057
1 0.526331535975138
1 0.513104804356893
1 0.528877401351929
1 0.741244347890218
1 0.955702527364095
1 0.934113963445028
1 0.781034072240194
1 0.757644939422607
1 0.744977696736654
1 0.747193193435669
1 0.78089648882548
};
\addplot [black, forget plot]
table {%
2 1.12533238728841
2 0.958236535390218
};
\addplot [black, forget plot]
table {%
2 1.23680621782939
2 1.39837025006612
};
\addplot [black, forget plot]
table {%
1.875 0.958236535390218
2.125 0.958236535390218
};
\addplot [black, forget plot]
table {%
1.875 1.39837025006612
2.125 1.39837025006612
};
\addplot [black, mark=*, mark size=2, mark options={solid,fill opacity=0}, only marks, forget plot]
table {%
2 0.95440452893575
2 0.953546714782715
2 0.936383930842082
2 0.939684025446574
2 0.956252892812093
2 0.930275789896647
2 1.41940938631694
2 1.45192476908366
2 1.42898216247559
2 1.49729469617208
};
\addplot [black, forget plot]
table {%
4 0.996147441864014
4 0.652026144663493
};
\addplot [black, forget plot]
table {%
4 1.23237223625183
4 1.58598256111145
};
\addplot [black, forget plot]
table {%
3.875 0.652026144663493
4.125 0.652026144663493
};
\addplot [black, forget plot]
table {%
3.875 1.58598256111145
4.125 1.58598256111145
};
\addplot [black, mark=*, mark size=2, mark options={solid,fill opacity=0}, only marks, forget plot]
table {%
4 0.623666715621948
4 0.618832365671794
4 0.570517269770304
4 0.526794322331746
4 0.617969369888306
4 0.547071234385173
4 0.559671910603841
4 0.638786697387695
4 0.633690452575684
4 0.629054880142212
4 0.517039664586385
4 1.61602778434753
4 1.7409605662028
4 1.5869257291158
4 1.61655958493551
4 1.63210299809774
4 1.61206223169963
};
\addplot [black, forget plot]
table {%
5 2.26274094581604
5 1.50017568270365
};
\addplot [black, forget plot]
table {%
5 2.77903975248337
5 3.55219677289327
};
\addplot [black, forget plot]
table {%
4.875 1.50017568270365
5.125 1.50017568270365
};
\addplot [black, forget plot]
table {%
4.875 3.55219677289327
5.125 3.55219677289327
};
\addplot [black, mark=*, mark size=2, mark options={solid,fill opacity=0}, only marks, forget plot]
table {%
5 1.45592495600382
5 1.40764393806458
5 1.2833592414856
5 1.35416332880656
5 1.45108742713928
5 1.25394150416056
5 1.28389678001404
5 1.46450239817301
5 1.48458580970764
5 1.44861194292704
5 1.20047470728556
5 4.03821438153585
5 3.66184171040853
5 3.59250340461731
5 3.5944111029307
};
\path [draw=black, fill=color1]
(axis cs:0.75,0.615392239888509)
--(axis cs:1.25,0.615392239888509)
--(axis cs:1.25,0.664434957504273)
--(axis cs:0.75,0.664434957504273)
--(axis cs:0.75,0.615392239888509)
--cycle;
\path [draw=black, fill=color2]
(axis cs:1.75,1.12533238728841)
--(axis cs:2.25,1.12533238728841)
--(axis cs:2.25,1.23680621782939)
--(axis cs:1.75,1.23680621782939)
--(axis cs:1.75,1.12533238728841)
--cycle;
\path [draw=black, fill=color1]
(axis cs:3.75,0.996147441864014)
--(axis cs:4.25,0.996147441864014)
--(axis cs:4.25,1.23237223625183)
--(axis cs:3.75,1.23237223625183)
--(axis cs:3.75,0.996147441864014)
--cycle;
\path [draw=black, fill=color2]
(axis cs:4.75,2.26274094581604)
--(axis cs:5.25,2.26274094581604)
--(axis cs:5.25,2.77903975248337)
--(axis cs:4.75,2.77903975248337)
--(axis cs:4.75,2.26274094581604)
--cycle;
\addplot [color0, forget plot]
table {%
0.75 0.640124877293905
1.25 0.640124877293905
};
\addplot [color0, forget plot]
table {%
1.75 1.17633088429769
2.25 1.17633088429769
};
\addplot [color0, forget plot]
table {%
3.75 1.11822168827057
4.25 1.11822168827057
};
\addplot [color0, forget plot]
table {%
4.75 2.5357386747996
5.25 2.5357386747996
};

\addplot[area legend, line width=0.5pt, draw=black, fill=color1]
table[row sep=crcr] {%
x y\\
2.9375  3.7438966765\\
}--cycle;
\addlegendentry{Distributed}

\addplot[area legend, line width=0.5pt, draw=black, fill=color2]
table[row sep=crcr] {%
x y\\
2.6875  1.738208592\\
}--cycle;
\addlegendentry{Centralized}
\end{axis}

\end{tikzpicture}}
    \caption{The boxplots of the average computation time in for all the 1000 simulations.} \label{fig:boxplot_timer}
    \vspace{-5pt}
\end{figure}
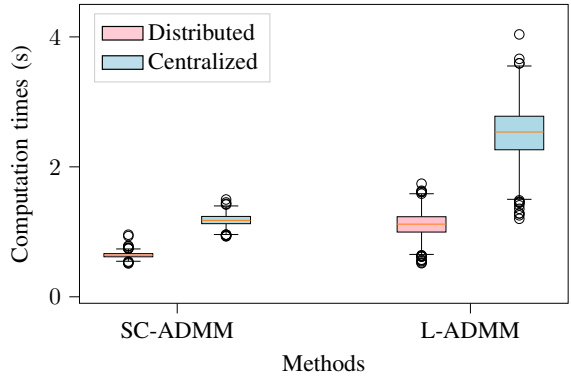

We now summarize all the results for the 3 different validation metrics obtained by both the algorithms in all the 1000 experiments and present them in the box plot format in Fig. \ref{fig:boxplotmetrics}. 
Similar to the aforementioned discussion for a single experiment example, from Fig. \ref{fig:boxplotmetrics} overall difference in the prediction results obtained by the L-ADMM algorithm and our proposed SC-ADMM approach is trivial although it is noticed that the predicted spatial field is highly comparable to the ground truth. 
Thus, in practice, for the purpose of prediction accuracy, it is suggested to solve the adaptive sampling optimization problem in a MRSN for environmental monitoring applications by the either L-ADMM or SC-ADMM techniques. 
Nonetheless, for practicality, we investigated computational complexity of both the algorithms by summarizing their computing time in all the 1000 experiments. 
The summarized computation time is shown by the boxplots in Fig. \ref{fig:boxplot_timer}, where we considered both the scenarios of using the distributed and centralized optimization. 
It is noted that in the centralized scenario all the computation was conducted at the central station while in the distributed scenario each individual robot calculated its own nonholonomic dynamics, control and movement constraints before sending the results to the central station, as presented in Section \ref{sec:pxadmm}. Fig. \ref{fig:boxplot_timer} obviously shows that the distributed paradigm is computationally better than the centralized model and that the proposed SC-ADMM algorithm runs much faster than the L-ADMM counterpart given the same system set-up. 
That is, the highly computational efficiency of the SC-ADMM approach leads to preferences in using the proposed algorithm to practical implementation in real-time systems.

\begin{figure*}[!tb]
\centering
\begin{subfigure}{.32\textwidth}
\centering
\scalebox{0.6}{\begin{tikzpicture}

\definecolor{color0}{rgb}{0.12156862745098,0.466666666666667,0.705882352941177}
\definecolor{color1}{rgb}{1,0.498039215686275,0.0549019607843137}

\begin{axis}[
width=8.4cm,
height=6.0cm,
legend cell align={left},
legend style={fill opacity=0.8, draw opacity=1, text opacity=1, draw=white!80!black},
tick align=outside,
tick pos=left,
x grid style={white!69.0196078431373!black},
xlabel={Time step},
xmajorgrids,
xmin=-0.2, xmax=15.2,
xtick={0,1,2,3,4,5,6,7,8,9,10,11,12,13,14,15},
xtick style={color=black},
y grid style={white!69.0196078431373!black},
ylabel={ALPV},
ymajorgrids,
ymin=-10, ymax=-1,
ytick={-10,-8,-6,-4,-2},
ytick style={color=black}
]
\addplot [semithick, color0, mark=square*, mark size=2, mark options={solid}]
table {%
0 -1.96901529379818
1 -3.71377582961434
2 -3.99582417463942
3 -3.56131859927412
4 -3.93068576418982
5 -4.72846093661308
6 -4.81392099039951
7 -5.59659095969907
8 -6.4247864374636
9 -7.19171863182901
10 -7.18354645374263
11 -7.28574881895865
12 -7.55848209009449
13 -7.85788421219075
14 -7.87987861025115
15 -8.09543897567758
};
\addlegendentry{SC-ADMM}
\addplot [semithick, color1, mark=square*, mark size=2, mark options={solid}]
table {%
0 -1.96903105238917
1 -2.29051716029119
2 -3.51753847138146
3 -4.18369389789123
4 -3.93302558281057
5 -4.21363463429287
6 -4.79617063414125
7 -5.62333072608503
8 -6.03470156280548
9 -6.56538521964952
10 -6.46623983529812
11 -6.63489622526865
12 -6.86981257224462
13 -7.19151693185924
14 -7.57528950020906
15 -7.74857062634818
};
\addlegendentry{L-ADMM}
\end{axis}

\end{tikzpicture}}
\caption{The average logarithm of the variances} \label{fig:sumlog}
\end{subfigure}
\begin{subfigure}{.32\textwidth}
\centering
\scalebox{0.6}{\begin{tikzpicture}

\definecolor{color0}{rgb}{0.12156862745098,0.466666666666667,0.705882352941177}
\definecolor{color1}{rgb}{1,0.498039215686275,0.0549019607843137}

\begin{axis}[
width=8.8cm,
height=6.0cm,
legend cell align={left},
legend style={fill opacity=0.8, draw opacity=1, text opacity=1, draw=white!80!black},
tick align=outside,
tick pos=left,
x grid style={white!69.0196078431373!black},
xlabel={Time step},
xmajorgrids,
xmin=-0.2, xmax=15.2,
xtick={0,1,2,3,4,5,6,7,8,9,10,11,12,13,14,15},
xtick style={color=black},
y grid style={white!69.0196078431373!black},
ylabel={RMSE},
ymajorgrids,
ymin=-0.05, ymax=1.05,
ytick={0,0.2,0.4,0.6,0.8,1.0},
ytick style={color=black}
]
\addplot [semithick, color0, mark=square*, mark size=2, mark options={solid}]
table {%
0 0.726400572332009
1 0.828310700472082
2 0.763012106921547
3 0.533702644542037
4 0.450553226799308
5 0.46970593283422
6 0.485554711333039
7 0.134449616571676
8 0.0886911141583624
9 0.0434946118705557
10 0.0187969370720371
11 0.020757895472715
12 0.0172738014324018
13 0.0137230825533717
14 0.0127626672149608
15 0.0129165416579055
};
\addlegendentry{SC-ADMM}
\addplot [semithick, color1, mark=square*, mark size=2, mark options={solid}]
table {%
0 0.72639940218219
1 0.661171915581086
2 0.665966547840719
3 0.644576825788782
4 0.362180074194654
5 0.256288694469371
6 0.0940267312070629
7 0.0779009060472146
8 0.0748572466024566
9 0.107408304147938
10 0.0900315534733581
11 0.0910100644550018
12 0.096485805260663
13 0.0591566527555541
14 0.0280315614824268
15 0.0214501546153594
};
\addlegendentry{L-ADMM}
\end{axis}

\end{tikzpicture}}
\caption{The root mean squared errors} \label{fig:rmse}
\end{subfigure}
\begin{subfigure}{.32\textwidth}
\centering
\scalebox{0.6}{\begin{tikzpicture}

\definecolor{color0}{rgb}{0.12156862745098,0.466666666666667,0.705882352941177}
\definecolor{color1}{rgb}{1,0.498039215686275,0.0549019607843137}

\begin{axis}[
width=8.8cm,
height=6.0cm,
legend cell align={left},
legend style={fill opacity=0.8, draw opacity=1, text opacity=1, draw=white!80!black},
tick align=outside,
tick pos=left,
x grid style={white!69.0196078431373!black},
xlabel={Time step},
xmajorgrids,
xmin=-0.2, xmax=15.2,
xtick={0,1,2,3,4,5,6,7,8,9,10,11,12,13,14,15},
xtick style={color=black},
y grid style={white!69.0196078431373!black},
ylabel={MAE},
ymajorgrids,
ymin=-0.1, ymax=3.25,
ytick={0,0.5,1.0,1.5,2.0,2.5,3.0},
ytick style={color=black}
]
\addplot [semithick, color0, mark=square*, mark size=2, mark options={solid}]
table {%
0 2.83234622201741
1 2.97273822445814
2 2.65538265624792
3 2.39501877354126
4 2.18422906884486
5 2.26825587572603
6 1.61862292089901
7 0.424284389780485
8 0.413024871004858
9 0.193156531856488
10 0.0647987426532808
11 0.0935219354303207
12 0.073355015354565
13 0.0493600541970913
14 0.0433904064331117
15 0.0432558379726231
};
\addlegendentry{SC-ADMM}
\addplot [semithick, color1, mark=square*, mark size=2, mark options={solid}]
table {%
0 2.8323424614805
1 2.89499627174681
2 3.09606296154774
3 3.17595928015448
4 1.30350650853705
5 1.20609126791409
6 0.367986497456759
7 0.432586863411501
8 0.387463854032742
9 0.474962466223463
10 0.44770646484837
11 0.421787439378313
12 0.415492266606066
13 0.294274286092705
14 0.157595624008675
15 0.142692373543138
};
\addlegendentry{L-ADMM}
\end{axis}

\end{tikzpicture}}
\caption{The maximum absolute errors} \label{fig:mae}
\end{subfigure}
\caption{The validation metrics for an experiment.}
\label{fig:metrics}
\vspace{-5pt}
\end{figure*}
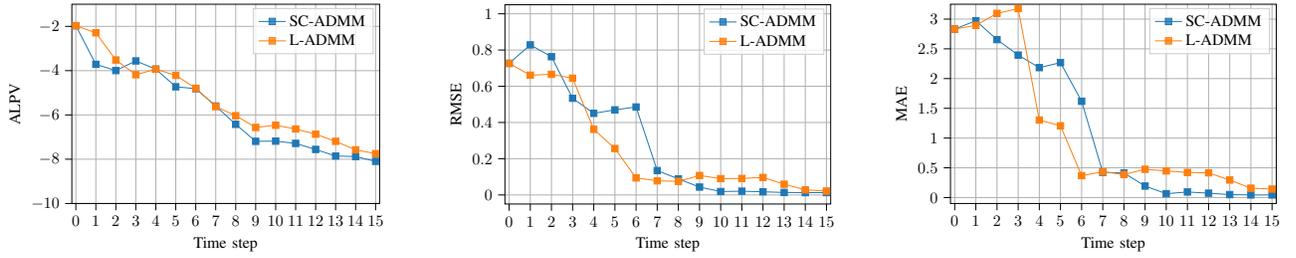

\begin{figure*}[!tb]
\centering
\begin{subfigure}{.32\textwidth}
\centering
\scalebox{.6}{\begin{tikzpicture}

\definecolor{color0}{rgb}{1,0.498039215686275,0.0549019607843137}
\definecolor{color1}{rgb}{1,0.752941176470588,0.796078431372549}
\definecolor{color2}{rgb}{0.67843137254902,0.847058823529412,0.901960784313726}

\begin{axis}[
width=8.8cm,
height=6.0cm,
tick align=outside,
tick pos=left,
x grid style={white!69.0196078431373!black},
xmin=0.5, xmax=4.5,
xtick style={color=black},
xlabel={Methods},
xmin=0.5, xmax=2.5,
xtick style={color=black},
xtick={1,2},
xticklabels={SC-ADMM,L-ADMM},
y grid style={white!69.0196078431373!black},
ylabel={ALPV},
ymin=-9.5, ymax=-5.0,
ytick style={color=black}
]
\addplot [black]
table {%
1 -7.97125935421588
1 -8.86947346256206
};
\addplot [black]
table {%
1 -7.35229819201124
1 -6.42514009930885
};
\addplot [black]
table {%
0.9625 -8.86947346256206
1.0375 -8.86947346256206
};
\addplot [black]
table {%
0.9625 -6.42514009930885
1.0375 -6.42514009930885
};
\addplot [black, mark=*, mark size=2, mark options={solid,fill opacity=0}, only marks]
table {%
1 -8.92268347154377
1 -8.97159790770883
1 -6.260387504295
1 -6.26521062384728
1 -6.40198621499612
1 -6.40447179266201
1 -6.10781725147017
1 -6.38569655982268
1 -6.24412905242009
1 -6.24812984835498
1 -5.66147636100912
1 -6.34728323192352
1 -6.19674134434911
1 -6.35064092614354
1 -6.01698066741748
1 -6.28733329129176
1 -4.81117037257965
1 -6.33520163959245
};
\addplot [black]
table {%
2 -7.85784970225548
2 -8.62501563048976
};
\addplot [black]
table {%
2 -7.30241783781631
2 -6.48805246791685
};
\addplot [black]
table {%
1.9625 -8.62501563048976
2.0375 -8.62501563048976
};
\addplot [black]
table {%
1.9625 -6.48805246791685
2.0375 -6.48805246791685
};
\addplot [black, mark=*, mark size=2, mark options={solid,fill opacity=0}, only marks]
table {%
2 -8.69946472842508
2 -8.84170277229111
2 -8.70234153474301
2 -6.2912066967612
2 -5.75441530832027
2 -6.35178373288216
2 -6.31531110503551
2 -6.30841859865209
2 -6.06473098237975
2 -5.70957347629954
2 -6.43762215546387
2 -6.10900823330407
2 -6.28542646605319
2 -6.12484048102789
2 -6.2464910661669
2 -6.19378383524859
2 -6.3081561295802
2 -6.40358436267032
2 -5.68352505993583
2 -6.16932415149609
2 -6.09487690104134
2 -6.37220319374967
2 -6.25504474059438
2 -6.39721614866033
2 -6.15857579075436
};
\path [draw=black, fill=color1]
(axis cs:0.925,-7.97125935421588)
--(axis cs:1.075,-7.97125935421588)
--(axis cs:1.075,-7.35229819201124)
--(axis cs:0.925,-7.35229819201124)
--(axis cs:0.925,-7.97125935421588)
--cycle;
\path [draw=black, fill=color2]
(axis cs:1.925,-7.85784970225548)
--(axis cs:2.075,-7.85784970225548)
--(axis cs:2.075,-7.30241783781631)
--(axis cs:1.925,-7.30241783781631)
--(axis cs:1.925,-7.85784970225548)
--cycle;
\addplot [color0]
table {%
0.925 -7.69142002555865
1.075 -7.69142002555865
};
\addplot [color0]
table {%
1.925 -7.59052916938491
2.075 -7.59052916938491
};
\end{axis}

\end{tikzpicture}}
\caption{The average logarithm of the variances} \label{fig:boxplot_logvar}
\end{subfigure}
\begin{subfigure}{.32\textwidth}
\centering
\scalebox{.6}{\begin{tikzpicture}

\definecolor{color0}{rgb}{1,0.498039215686275,0.0549019607843137}
\definecolor{color1}{rgb}{1,0.752941176470588,0.796078431372549}
\definecolor{color2}{rgb}{0.67843137254902,0.847058823529412,0.901960784313726}

\begin{axis}[
width=8.8cm,
height=6.0cm,
tick align=outside,
tick pos=left,
x grid style={white!69.0196078431373!black},
xmin=0.5, xmax=4.5,
xlabel={Methods},
xmin=0.5, xmax=2.5,
xtick style={color=black},
xtick={1,2},
xticklabels={SC-ADMM,L-ADMM},
y grid style={white!69.0196078431373!black},
ylabel={RMSEs},
ymin=-0.05, ymax=0.55,
ytick style={color=black}
]
\addplot [black]
table {%
1 0.028188110511047
1 0.0128131110152992
};
\addplot [black]
table {%
1 0.0733675932647505
1 0.14033333848229
};
\addplot [black]
table {%
0.9625 0.0128131110152992
1.0375 0.0128131110152992
};
\addplot [black]
table {%
0.9625 0.14033333848229
1.0375 0.14033333848229
};
\addplot [black, mark=*, mark size=2, mark options={solid,fill opacity=0}, only marks]
table {%
1 0.20404941195167
1 0.356057829305687
1 0.149737005171828
1 0.223669347543775
1 0.142281727010325
1 0.169977389350565
1 0.260574167780534
1 0.225398760926891
1 0.187107393533753
1 0.228181372589236
1 0.216755769806502
1 0.215947985623303
1 0.324610023045704
1 0.189022939614217
1 0.319958378219551
1 0.21267792633576
1 0.279852776662674
1 0.145406259438004
1 0.172388086297079
1 0.190383431303529
1 0.331313047153849
1 0.20764834553044
1 0.156971868997604
1 0.153760264569273
1 0.162378121094724
1 0.211469469978766
1 0.167740685347519
1 0.170268944496505
1 0.177731397658968
1 0.1853504466829
1 0.210955408368959
1 0.189724287984858
1 0.193387764060404
1 0.288428346880864
1 0.214126936447138
1 0.196473896310021
1 0.147261088888855
1 0.220964827514921
1 0.147856792693398
1 0.213276227391788
1 0.191125986597302
1 0.174018576829908
1 0.14173286300444
1 0.141296100541451
1 0.143153289409176
1 0.186499070287016
1 0.1941994013649
1 0.216745910851891
1 0.222541023832897
1 0.145422674731555
1 0.306588433870047
1 0.179631063748172
1 0.406984133855269
1 0.32656516202762
};
\addplot [black]
table {%
2 0.0220028449567967
2 0.0115610778055383
};
\addplot [black]
table {%
2 0.0613081018388208
2 0.120166077741409
};
\addplot [black]
table {%
1.9625 0.0115610778055383
2.0375 0.0115610778055383
};
\addplot [black]
table {%
1.9625 0.120166077741409
2.0375 0.120166077741409
};
\addplot [black, mark=*, mark size=2, mark options={solid,fill opacity=0}, only marks]
table {%
2 0.13900819367638
2 0.12084967481365
2 0.164677154361971
2 0.141289729339878
2 0.268691254255048
2 0.142124885645389
2 0.194002841094423
2 0.274440678647811
2 0.14338347710443
2 0.339387275761284
2 0.131589601315781
2 0.343247700000651
2 0.157999759511771
2 0.129789348728222
2 0.180081949795287
2 0.162555738744611
2 0.142393466830189
2 0.158162294259657
2 0.133877698289755
2 0.245468819696106
2 0.178874137065276
2 0.258471705489515
2 0.208502254198212
2 0.135346433276572
2 0.168557356881206
2 0.137831388719312
2 0.451595507043956
2 0.164049091979711
2 0.120792768858246
2 0.216606042818863
2 0.139266555897757
2 0.225780020164994
2 0.259916923568716
2 0.285487193854853
2 0.133364221097685
2 0.158685177363894
2 0.180144865102993
2 0.161166792768305
2 0.124589818978344
2 0.135608797079113
2 0.125422174642194
2 0.195346283727546
2 0.257222412783024
2 0.171952797605696
2 0.136842837533227
2 0.273333852795883
2 0.142802236883745
2 0.225656228730813
2 0.314455551353319
2 0.22161342188529
2 0.236869422822931
2 0.123082123182283
2 0.256854065089586
2 0.13916822274768
2 0.243229181912949
2 0.240932546251809
2 0.147722295992488
2 0.138657889339779
2 0.129008239989077
2 0.237185845306309
2 0.131509510918256
2 0.278511873763571
2 0.142353075609073
2 0.293558865483978
2 0.127004653058858
2 0.192780981104659
2 0.158930752136234
2 0.185899378713196
2 0.136034172818184
2 0.156370837608936
2 0.131182921446531
2 0.178554087703523
2 0.193725973620571
2 0.217790707405425
2 0.201459938401675
2 0.258339139379031
2 0.159310473793085
2 0.22703275130106
2 0.184790832408286
2 0.248597992693458
2 0.148304620588259
2 0.171723307078312
2 0.152014630992477
2 0.155019776051661
2 0.18881718332511
2 0.140984173913306
2 0.243934363053693
2 0.120617849257314
2 0.138980746051832
2 0.270999491180197
2 0.222193850957646
2 0.168886609910447
2 0.145639198049193
2 0.162361914801532
2 0.206958065691594
2 0.229832464134917
2 0.134088257335172
2 0.127573113397393
2 0.156220855822613
2 0.218050562569172
2 0.156604662202914
};
\path [draw=black, fill=color1]
(axis cs:0.925,0.028188110511047)
--(axis cs:1.075,0.028188110511047)
--(axis cs:1.075,0.0733675932647505)
--(axis cs:0.925,0.0733675932647505)
--(axis cs:0.925,0.028188110511047)
--cycle;
\path [draw=black, fill=color2]
(axis cs:1.925,0.0220028449567967)
--(axis cs:2.075,0.0220028449567967)
--(axis cs:2.075,0.0613081018388208)
--(axis cs:1.925,0.0613081018388208)
--(axis cs:1.925,0.0220028449567967)
--cycle;
\addplot [color0]
table {%
0.925 0.0424462640043545
1.075 0.0424462640043545
};
\addplot [color0]
table {%
1.925 0.0330308018576875
2.075 0.0330308018576875
};
\end{axis}

\end{tikzpicture}}
\caption{The root mean squared errors} \label{fig:boxplot_rmse}
\end{subfigure}
\begin{subfigure}{.32\textwidth}
\centering
\scalebox{0.6}{\begin{tikzpicture}

\definecolor{color0}{rgb}{1,0.498039215686275,0.0549019607843137}
\definecolor{color1}{rgb}{1,0.752941176470588,0.796078431372549}
\definecolor{color2}{rgb}{0.67843137254902,0.847058823529412,0.901960784313726}

\begin{axis}[
width=8.8cm,
height=6.0cm,
tick align=outside,
tick pos=left,
x grid style={white!69.0196078431373!black},
xmin=0.5, xmax=4.5,
xlabel={Methods},
xmin=0.5, xmax=2.5,
xtick style={color=black},
xtick={1,2},
xticklabels={SC-ADMM,L-ADMM},
y grid style={white!69.0196078431373!black},
ylabel={MAEs},
ymin=-0.05, ymax=2.5,
ytick style={color=black}
]
\addplot [black]
table {%
1 0.156499444457722
1 0.0469478187385235
};
\addplot [black]
table {%
1 0.480973191614233
1 0.95874178840792
};
\addplot [black]
table {%
0.9625 0.0469478187385235
1.0375 0.0469478187385235
};
\addplot [black]
table {%
0.9625 0.95874178840792
1.0375 0.95874178840792
};
\addplot [black, mark=*, mark size=2, mark options={solid,fill opacity=0}, only marks]
table {%
1 1.08202259220814
1 1.96904271077693
1 1.24352788446655
1 1.74824330782826
1 1.06734236937983
1 1.58402998485871
1 1.32814955780986
1 1.37812910480786
1 1.429190632627
1 1.39364781503814
1 1.27237414930816
1 2.08813016132386
1 1.68376538945127
1 1.06080852684864
1 1.1602851597815
1 1.23703263941409
1 1.3480747630689
1 1.4830162273531
1 1.98138012654926
1 1.60483568117738
1 1.29277304265438
1 1.30500822475629
1 1.23233634068038
1 1.31063677999318
1 1.38844850606724
1 1.36390504326789
1 1.34171193412715
1 1.57112279370707
1 1.70795616701241
1 1.39830651549267
1 1.31731857595505
1 1.06324864419245
1 1.11635859525178
1 1.55955421164399
1 1.19933907441127
1 1.51078490648316
1 1.31865776920634
1 1.15708410471475
1 1.00367255333931
1 1.19867906054202
1 1.12589513182198
1 1.31805156463597
1 1.34284593123388
1 1.0051953382423
1 1.10511705606191
1 2.00795367071355
1 1.26257761552022
1 1.60755574688958
1 0.985216576166223
1 1.76204977244038
};
\addplot [black]
table {%
2 0.115486521129727
2 0.041543595914213
};
\addplot [black]
table {%
2 0.417948514453962
2 0.867382485173462
};
\addplot [black]
table {%
1.9625 0.041543595914213
2.0375 0.041543595914213
};
\addplot [black]
table {%
1.9625 0.867382485173462
2.0375 0.867382485173462
};
\addplot [black, mark=*, mark size=2, mark options={solid,fill opacity=0}, only marks]
table {%
2 1.02729900103294
2 0.915654044100449
2 1.28934003719854
2 1.01139478041135
2 1.72028506287422
2 0.988797638938063
2 0.937083039446414
2 1.18610543149483
2 1.21758651432337
2 1.73495448293987
2 1.6201523106406
2 1.43437976918553
2 0.994415173035062
2 1.10461096827525
2 1.2596935303204
2 1.07586033339107
2 1.70056313031935
2 1.37259292265995
2 1.30168549165137
2 1.41400622002449
2 1.1713543382077
2 1.65467124246083
2 1.21259277226334
2 0.959815124093399
2 1.49871877325172
2 1.36621093235198
2 1.61362812032888
2 1.84325430179629
2 0.900330970516791
2 1.23890785893371
2 0.915327580261984
2 1.26764018374272
2 0.99901883691259
2 1.0078612637359
2 0.965719150046663
2 1.63841195724402
2 0.883242857199431
2 1.05364817937686
2 1.62645257511634
2 1.00770986428402
2 2.01346450424467
2 1.58378892831717
2 1.26076775878081
2 1.30987206748743
2 1.84930026677113
2 1.40880665660113
2 0.921829402645621
2 1.24226278046063
2 1.15346289913332
2 1.78576484591051
2 1.74131269422848
2 0.973551791907262
2 1.12013812251475
2 0.97411829681975
2 1.08152717750011
2 0.878896632345381
2 1.28936149542952
2 1.47064094653848
2 1.12993093389881
2 1.31573400914361
2 1.16751211934458
2 1.31914511441132
2 1.66261342176392
2 1.56644641653996
2 1.05502001561112
2 1.70957498227418
2 1.19327833990042
2 0.906018628992928
2 1.18244876842819
2 1.18670655590192
2 1.02597158789153
2 1.72746360895509
2 0.907686341628338
2 1.26775853123287
2 1.13176392257487
2 1.00099893961578
2 1.08605407006606
2 1.09922720191904
2 1.40883805217343
2 1.52307570056222
2 0.984999393894739
2 1.15601038092678
2 1.17300795783727
2 1.2906014582239
2 0.882724103559006
};
\path [draw=black, fill=color1]
(axis cs:0.925,0.156499444457722)
--(axis cs:1.075,0.156499444457722)
--(axis cs:1.075,0.480973191614233)
--(axis cs:0.925,0.480973191614233)
--(axis cs:0.925,0.156499444457722)
--cycle;
\path [draw=black, fill=color2]
(axis cs:1.925,0.115486521129727)
--(axis cs:2.075,0.115486521129727)
--(axis cs:2.075,0.417948514453962)
--(axis cs:1.925,0.417948514453962)
--(axis cs:1.925,0.115486521129727)
--cycle;
\addplot [color0]
table {%
0.925 0.275450528142777
1.075 0.275450528142777
};
\addplot [color0]
table {%
1.925 0.212285053444774
2.075 0.212285053444774
};
\end{axis}

\end{tikzpicture}}
\caption{The maximum absolute errors} \label{fig:boxplot_mae}
\end{subfigure}
\caption{The box plots of the validation metrics at the $15^{th}$ time step for all the 1000 experiments.}
\label{fig:boxplotmetrics}
\vspace{-10pt}
\end{figure*}
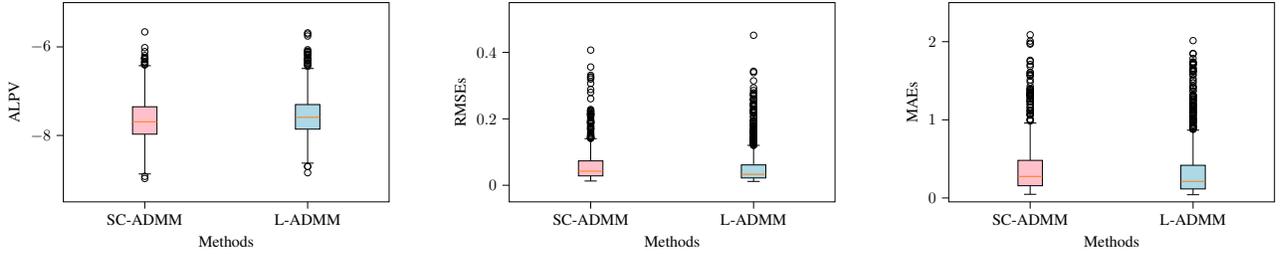

\section{Conclusions}
\label{sec:conclusion}

The paper has presented a discussion about the adaptive sampling in a nonholonomic MRSN for effectively monitoring an environmental spatial field. 
To the best of our knowledge, it is the first time the paper has taken all the control, movement and nonholonomic dynamic constraints of the mobile sensors into consideration, which makes the sampling optimization problem highly nonlinear, nonconvex and complex. 
We have first discussed how to solve the optimization problem by the L-ADMM algorithm and discovered the computational complexity in the algorithm due to its indicator function definition. 
We have then proposed a novel but efficient SC-ADMM approach that exploits the first-order approximation to tractably handle non-convexity and high complexity of the objective function and the successive convexification programming to sequentially convexify the nonlinear dynamic constraints. 
Thus, the proposed SC-ADMM approach can computationally effectively and accurately address the adaptive sampling optimization problem as compared with the L-ADMM counterpart. 
We have implemented both the L-ADMM and SC-ADMM algorithms in the 1000 experiments in a synthetic environment using the realistic indoor temperature dataset. 
Though the obtained results demonstrate that both the algorithms can provide accurate prediction of the spatial temperature field at the unobserved locations, the SC-ADMM method run much faster, which promises its potential practicality.

\hilitediff{In the future work, we will implement the proposed approach in a realistic mobile sensor system to verify its practical effectiveness.}
\vspace{-10pt}

\bibliographystyle{IEEEtran}
\bibliography{IEEEabrv,references}

\end{document}